\documentclass{article}

\newif\ifanonymize
\anonymizefalse

\newif\ifarxiv
\arxivtrue

\PassOptionsToPackage{numbers, sort}{natbib}

\ifanonymize
    \usepackage{nips_2016}
\else
    \usepackage[final]{nips_2016}
\fi

\usepackage[utf8]{inputenc} 
\usepackage[T1]{fontenc}    
\usepackage{url}            
\usepackage{booktabs}       
\usepackage{amsmath, amsfonts}       
\usepackage{amsthm}
\usepackage{nicefrac}       
\usepackage{microtype}      
\usepackage{bm}
\usepackage{xr}
\usepackage[algoruled, noline]{algorithm2e}
\usepackage{graphicx}
\usepackage{subfig}
\usepackage{enumitem}
\usepackage{hyperref}

\newtheorem{proposition}{Proposition}

\newcommand{\vect}[1]{\mathbf{#1}}
\newcommand{\mat}[1]{\mathbf{#1}}
\newcommand{\br}[1]{\mathopen{}\left(#1\right)\mathclose{}}
\newcommand{\set}[1]{\left\{#1\right\}}
\newcommand{\pair}[2]{\br{#1,#2}}

\newcommand{\norm}[1]{\left\|#1\right\|}

\newcommand{\prob}[1]{p\br{#1}}
\newcommand{\hatprob}[1]{\hat{p}\br{#1}}
\newcommand{\g}{\,|\,}
\newcommand{\kl}[2]{D_{\mathrm{KL}}\br{#1\,\|\,#2}}
\newcommand{\avg}[2]{\left\langle#1\right\rangle_{#2}}

\newcommand{\gaussian}[3]{\mathcal{N}\br{#1\g #2,#3}}
\newcommand{\uniform}[3]{\mathcal{U}\br{#1\g #2,#3}}

\newcommand{\algref}[1]{\hyperlink{#1_anchor}{\ref*{#1}}}

\newcommand{\supref}{\ref*}

\newcommand{\ourtitle}{Fast $\epsilon$-free Inference of Simulation Models with Bayesian Conditional Density Estimation}

\author{
George Papamakarios\\
School of Informatics\\
University of Edinburgh\\
\texttt{g.papamakarios@ed.ac.uk}\\
\And
Iain Murray\\
School of Informatics\\
University of Edinburgh\\
\texttt{i.murray@ed.ac.uk}\\
}

\title{\ourtitle}

\begin{document}

\maketitle

\begin{abstract}
Many statistical models can be simulated forwards but have intractable likelihoods. Approximate Bayesian Computation (ABC) methods are used to infer properties of these models from data. Traditionally these methods approximate the posterior over parameters by conditioning on data being inside an $\epsilon$-ball around the observed data, which is only correct in the limit $\epsilon\!\rightarrow\!0$. Monte Carlo methods can then draw samples from the approximate posterior to approximate predictions or error bars on parameters. These algorithms critically slow down as $\epsilon\!\rightarrow\!0$, and in practice draw samples from a broader distribution than the posterior.
We propose a new approach to likelihood-free inference based on Bayesian conditional density estimation. Preliminary inferences based on limited simulation data are used to guide later simulations. In some cases, learning an accurate parametric representation of the entire true posterior distribution requires fewer model simulations than Monte Carlo ABC methods need to produce a single sample from an approximate posterior.
\end{abstract}

\section{Introduction}
\label{sec:introduction}

A simulator-based model is a data-generating process described by a computer program, usually with some free parameters we need to learn from data. Simulator-based modelling lends itself naturally to scientific domains such as evolutionary biology \citep{Beaumont:2002}, ecology \citep{Wood:2010}, disease epidemics~\citep{Gutmann:2015}, economics~\citep{Gourieroux:1993} and cosmology \citep{Schafer:2012}, where observations are best understood as products of underlying physical processes. Inference in these models amounts to discovering plausible parameter settings that could have generated our observed data. The application domains mentioned can require properly calibrated distributions that express uncertainty over plausible parameters, rather than just point estimates, in order to reach scientific conclusions or make decisions.

As an analytical expression for the likelihood of parameters given observations is typically not available for simulator-based models, conventional likelihood-based Bayesian inference is not applicable. An alternative family of algorithms for likelihood-free inference has been developed, referred to as Approximate Bayesian Computation (ABC)\@. These algorithms simulate the model repeatedly and only accept parameter settings which generate synthetic data similar to the observed data, typically gathered in a real-world experiment.

Rejection ABC \citep{Pritchard:1999}, the most basic ABC algorithm, simulates the model for each setting of proposed parameters, and rejects parameters if the generated data is not within a certain distance from the observations. The accepted parameters form a set of independent samples from an approximate posterior. Markov Chain Monte Carlo ABC (MCMC-ABC) \citep{Marjoram:2003} is an improvement over rejection ABC which, instead of independently proposing parameters, explores the parameter space by perturbing the most recently accepted parameters. Sequential Monte Carlo ABC (SMC-ABC) \citep{Beaumont:2009, Bonassi:2015} uses importance sampling to simulate a sequence of slowly-changing distributions, the last of which is an approximation to the parameter posterior.

Conventional ABC algorithms such as the above suffer from three drawbacks. First, they only represent the parameter posterior as a set of (possibly weighted or correlated) samples. A sample-based representation easily gives estimates and error bars of individual parameters, and model predictions. However these computations are noisy, and it is not obvious how to perform some other computations using samples, such as combining posteriors from two separate analyses. Second, the parameter samples do not come from the correct Bayesian posterior, but from an approximation based on assuming a pseudo-observation that the data is within an $\epsilon$-ball centred on the data actually observed. Third, as the $\epsilon$-tolerance is reduced, it can become impractical to simulate the model enough times to match the observed data even once. When simulations are expensive to perform, good quality inference becomes impractical.

We propose a parametric approach to likelihood-free inference, which unlike conventional ABC does not suffer from the above three issues. Instead of returning samples from an $\epsilon$-approximation to the posterior, our approach learns a parametric approximation to the exact posterior, which can be made as accurate as required. Preliminary fits to the posterior are used to guide future simulations, which can reduce the number of simulations required to learn an accurate approximation by orders of magnitude. Our approach uses conditional density estimation with Bayesian neural networks, and draws upon advances in parametric density estimation, stochastic variational inference, and recognition networks, as discussed in the related work section.

\section{Bayesian conditional density estimation for likelihood-free inference}
\label{sec:neural_density_estimation}

\subsection{Simulator-based models and ABC}

Let $\bm{\theta}$ be a vector of parameters controlling a simulator-based model, and let $\vect{x}$ be a data vector generated by the model. The model may be provided as a probabilistic program that can be easily simulated, and implicitly defines a likelihood $\prob{\vect{x}\g\bm{\theta}}$, which we assume we cannot evaluate. Let $\prob{\bm{\theta}}$ encode our prior beliefs about the parameters. Given an observation $\vect{x}_o$, we are interested in the parameter posterior $\prob{\bm{\theta}\g \vect{x}=\vect{x}_o} \propto \prob{\vect{x}=\vect{x}_o\g \bm{\theta}}\, \prob{\bm{\theta}}$.

As the likelihood $\prob{\vect{x}=\vect{x}_o\g \bm{\theta}}$ is unavailable, conventional Bayesian inference cannot be carried out. The principle behind ABC is to approximate $\prob{\vect{x}=\vect{x}_o\g \bm{\theta}}$ by $\prob{\norm{\vect{x}-\vect{x}_o}<\epsilon\g \bm{\theta}}$ for a sufficiently small value of $\epsilon$, and then estimate the latter---e.g.~by Monte Carlo---using simulations from the model. Hence, ABC approximates the posterior by $\prob{\bm{\theta}\g \norm{\vect{x}-\vect{x}_o}<\epsilon}$, which is typically broader and more uncertain. ABC can trade off computation for accuracy by decreasing $\epsilon$, which improves the approximation to the posterior but requires more simulations from the model. However, the approximation becomes exact only when $\epsilon\rightarrow 0$, in which case simulations never match the observations, $\prob{\norm{\vect{x}-\vect{x}_o}<\epsilon\g \bm{\theta}}\rightarrow 0$, and existing methods break down. In this paper, we refer to $\prob{\bm{\theta}\g \vect{x}=\vect{x}_o}$ as the \emph{exact posterior}, as it corresponds to setting $\epsilon=0$ in $\prob{\bm{\theta}\g \norm{\vect{x}-\vect{x}_o}<\epsilon}$.

In most practical applications of ABC, $\vect{x}$ is taken to be a fixed-length vector of summary statistics that is calculated from data generated by the simulator, rather than the raw data itself. Extracting statistics is often necessary in practice, to reduce the dimensionality of the data and maintain $\prob{\norm{\vect{x}-\vect{x}_o}<\epsilon\g \bm{\theta}}$ to practically acceptable levels. For the purposes of this paper, we will make no distinction between raw data and summary statistics, and we will regard the calculation of summary statistics as part of the data generating process.

\subsection{Learning the posterior}

Rather than using simulations from the model in order to estimate an approximate likelihood, $\prob{\norm{\vect{x}-\vect{x}_o}<\epsilon\g \bm{\theta}}$, we will use the simulations to directly estimate $\prob{\bm{\theta}\g \vect{x}=\vect{x}_o}$. We will run simulations for parameters drawn from a distribution, $\tilde{p}\br{\bm{\theta}}$, which we shall refer to as the \emph{proposal prior}. The proposition below indicates how we can then form a consistent estimate of the exact posterior, using a flexible family of conditional densities, $q_{\bm{\phi}}\br{\bm{\theta}\g\vect{x}}$, parameterized by a vector $\bm{\phi}$.
\begin{proposition}\label{proposition}
We assume that each of a set of $N$ pairs $\pair{\bm{\theta}_n}{\vect{x}_n}$ was independently generated by
\begin{equation}
\bm{\theta}_n \sim \tilde{p}\br{\bm{\theta}}\quad\text{ and }\quad\vect{x}_n \sim \prob{\vect{x}\g \bm{\theta}_n}.
\end{equation}
In the limit $N\rightarrow\infty$, the probability of the parameter vectors $\prod_n q_{\bm{\phi}}(\bm{\theta}_n\g \vect{x}_n)$ is maximized w.r.t.~$\bm{\phi}$ if and only if
\begin{equation}
q_{\bm{\phi}}\br{\bm{\theta}\g\vect{x}} \propto \frac{\tilde{p}\br{\bm{\theta}}}{\prob{\bm{\theta}}}\,\prob{\bm{\theta}\g \vect{x}},
\label{eq:optimal_density_estimator}
\end{equation}
provided a setting of $\bm{\phi}$ that makes $q_{\bm{\phi}}\br{\bm{\theta}\g\vect{x}}$ proportional to $\frac{\tilde{p}\br{\bm{\theta}}}{\prob{\bm{\theta}}}\,\prob{\bm{\theta}\g \vect{x}}$ exists.
\end{proposition}
\emph{Intuition}: if we simulated enough parameters from the prior, the density estimator $q_{\bm{\phi}}$ would learn a conditional of the joint prior model over parameters and data, which is the posterior $\prob{\bm{\theta}\g \vect{x}}$. If we simulate parameters drawn from another distribution, we need to ``importance reweight'' the result. A more detailed proof can be found in 
\ifarxiv
Appendix~\ref{sec:proof}.
\else
Section~\supref{suppl-sec:proof} of the supplementary material.
\fi

The proposition above suggests the following procedure for learning the posterior: (a) propose a set of parameter vectors $\set{\bm{\theta}_n}$ from the proposal prior; (b) for each $\bm{\theta}_n$ run the simulator to obtain a corresponding data vector $\vect{x}_n$; (c) train $q_{\bm{\phi}}$ with maximum likelihood on $\set{\bm{\theta}_n,\vect{x}_n}$; and (d) estimate the posterior by
\begin{equation}
\hatprob{\bm{\theta}\g \vect{x}=\vect{x}_o} \propto\frac{\prob{\bm{\theta}}}{\tilde{p}\br{\bm{\theta}}}\,q_{\bm{\phi}}\br{\bm{\theta}\g\vect{x}_o}.
\label{eq:posterior_estimator}
\end{equation}
This procedure is summarized in Algorithm~\algref{alg:train_posterior}.

\subsection{Choice of conditional density estimator and proposal prior}

In choosing the types of density estimator $q_{\bm{\phi}}\br{\bm{\theta}\g\vect{x}}$ and proposal prior $\tilde{p}\br{\bm{\theta}}$, we need to meet the following criteria: (a) $q_{\bm{\phi}}$ should be flexible enough to represent the posterior but easy to train with maximum likelihood; (b) $\tilde{p}\br{\bm{\theta}}$ should be easy to evaluate and sample from; and (c) the right-hand side expression in Equation~\eqref{eq:posterior_estimator} should be easily evaluated and normalized.

We draw upon work on conditional neural density estimation and take $q_{\bm{\phi}}$ to be a Mixture Density Network (MDN) \citep{Bishop:1994} with fully parameterized covariance matrices. That is, $q_{\bm{\phi}}$ takes the form of a mixture of $K$ Gaussian components
$q_{\bm{\phi}}\br{\bm{\theta}\g\vect{x}} = \sum_k{\alpha_k\,\gaussian{\bm{\theta}}{\vect{m}_k}{\mat{S}_k}}$,
whose mixing coefficients $\set{\alpha_k}$, means $\set{\vect{m}_k}$ and covariance matrices $\set{\mat{S}_k}$ are computed by a feedforward neural network parameterized by $\bm{\phi}$, taking $\vect{x}$ as input. Such an architecture is capable of representing any conditional distribution arbitrarily accurately---provided the number of components $K$ and number of hidden units in the neural network are sufficiently large---while remaining trainable by backpropagation. The parameterization of the MDN is detailed in
\ifarxiv
Appendix~\ref{sec:mdn_parameterization}.
\else
Section~\supref{suppl-sec:mdn_parameterization} of the supplementary material.
\fi

We take the proposal prior to be a single Gaussian $\tilde{p}\br{\bm{\theta}} = \gaussian{\bm{\theta}}{\vect{m}_0}{\mat{S}_0}$, with mean $\vect{m}_0$ and full covariance matrix $\mat{S}_0$. Assuming the prior $\prob{\bm{\theta}}$ is a simple distribution (uniform or Gaussian, as is typically the case in practice), then this choice allows us to calculate $\hat{p}\br{\bm{\theta}\g \vect{x}=\vect{x}_o}$ in Equation~\eqref{eq:posterior_estimator} analytically. That is, $\hatprob{\bm{\theta}\g \vect{x}=\vect{x}_o}$ will be a mixture of $K$ Gaussians, whose parameters will be a function of $\set{\alpha_k, \vect{m}_k, \mat{S}_k}$ evaluated at $\vect{x}_o$ (as detailed in 
\ifarxiv
Appendix~\ref{sec:mog_calc_posterior}).
\else
Section~\supref{suppl-sec:mog_calc_posterior} of the supplementary material).
\fi

\subsection{Learning the proposal prior}

Simple rejection ABC is inefficient because the posterior $\prob{\bm{\theta}\g \vect{x}=\vect{x}_o}$ is typically much narrower than the prior $\prob{\bm{\theta}}$. A parameter vector $\bm{\theta}$ sampled from $\prob{\bm{\theta}}$ will rarely be plausible under $\prob{\bm{\theta}\g \vect{x}=\vect{x}_o}$ and will most likely be rejected. Practical ABC algorithms attempt to reduce the number of rejections by modifying the way they propose parameters; for instance, MCMC-ABC and SMC-ABC propose new parameters by perturbing parameters they already consider plausible, in the hope that nearby parameters remain plausible.

In our framework, the key to efficient use of simulations lies in the choice of proposal prior. If we take $\tilde{p}\br{\bm{\theta}}$ to be the actual prior, then $q_{\bm{\phi}}\br{\bm{\theta}\g\vect{x}}$ will learn the posterior for all $\vect{x}$, as can be seen from Equation~\eqref{eq:optimal_density_estimator}. Such a strategy however is grossly inefficient if we are only interested in the posterior for $\vect{x}=\vect{x}_o$. Conversely, if $\tilde{p}\br{\bm{\theta}}$ closely matches $\prob{\bm{\theta}\g \vect{x}=\vect{x}_o}$, then most simulations will produce samples that are highly informative in learning $q_{\bm{\phi}}\br{\bm{\theta}\g\vect{x}}$ for $\vect{x}=\vect{x}_o$. In other words, if we already knew the true posterior, we could use it to construct an efficient proposal prior for learning it.

We exploit this idea to set up a fixed-point system. Our strategy becomes to learn an efficient proposal prior that closely approximates the posterior as follows: (a)~initially take $\tilde{p}\br{\bm{\theta}}$ to be the prior $\prob{\bm{\theta}}$; (b)~propose $N$ samples $\set{\bm{\theta}_n}$ from $\tilde{p}\br{\bm{\theta}}$ and corresponding samples $\set{\vect{x}_n}$ from the simulator, and train $q_{\bm{\phi}}\br{\bm{\theta}\g\vect{x}}$ on them; (c)~approximate the posterior using Equation~\eqref{eq:posterior_estimator} and set $\tilde{p}\br{\bm{\theta}}$ to it; (d)~repeat until $\tilde{p}\br{\bm{\theta}}$ has converged. This procedure is summarized in Algorithm~\algref{alg:train_prop_prior}.

In the procedure above, as long as $q_{\bm{\phi}}\br{\bm{\theta}\g\vect{x}}$ has only one Gaussian component ($K=1$) then $\tilde{p}\br{\bm{\theta}}$ remains a single Gaussian throughout. Moreover, in each iteration we initialize $q_{\bm{\phi}}$ with the density estimator learnt in the iteration before, thus we keep training $q_{\bm{\phi}}$ throughout. This initialization allows us to use a small sample size $N$ in each iteration, thus making efficient use of simulations.

As we shall demonstrate in Section~\ref{sec:experiments}, the procedure above learns Gaussian approximations to the true posterior fast: in our experiments typically $4$--$6$ iterations of $200$--$500$ samples each were sufficient. This Gaussian approximation can be used as a rough but cheap approximation to the true posterior, or it can serve as a good proposal prior in Algorithm~\algref{alg:train_posterior} for efficiently fine-tuning a non-Gaussian multi-component posterior. If the second strategy is adopted, then we can reuse the single-component neural density estimator learnt in Algorithm~\algref{alg:train_prop_prior} to initialize $q_{\bm{\phi}}$ in Algorithm~\algref{alg:train_posterior}. The weights in the final layer of the MDN are replicated $K$ times, with small random perturbations to break symmetry.

\begin{figure}[t]
\begin{tabular}{cc}
\begin{minipage}{0.47\textwidth}
\hypertarget{alg:train_prop_prior_anchor}{}%
\begin{algorithm}[H]
initialize $q_{\bm{\phi}}\br{\bm{\theta}\g\vect{x}}$ with one component \\
$\tilde{p}\br{\bm{\theta}} \leftarrow \prob{\bm{\theta}}$ \\
\Repeat{$\tilde{p}\br{\bm{\theta}}$ has converged}{
\For{$n=1..N$}{
sample $\bm{\theta}_n \sim \tilde{p}\br{\bm{\theta}}$ \\
sample $\vect{x}_n \sim \prob{\vect{x}\g \bm{\theta}_n}$ \\
}
retrain $q_{\bm{\phi}}\br{\bm{\theta}\g\vect{x}}$ on $\set{\bm{\theta}_n, \vect{x}_n}$\\
$\tilde{p}\br{\bm{\theta}} \leftarrow \frac{\prob{\bm{\theta}}}{\tilde{p}\br{\bm{\theta}}}\,q_{\bm{\phi}}\br{\bm{\theta}\g\vect{x}_o}$
}
\caption{Training of proposal prior}
\label{alg:train_prop_prior}
\end{algorithm}
\end{minipage}
&
\begin{minipage}{0.47\textwidth}
\hypertarget{alg:train_posterior_anchor}{}%
\begin{algorithm}[H]
initialize $q_{\bm{\phi}}\br{\bm{\theta}\g\vect{x}}$ with K components \\
\tcp{if $q_{\bm{\phi}}$ available by Algorithm~\ref*{alg:train_prop_prior}}
\tcp{initialize by replicating its}
\tcp{one component $K$ times}
\For{$n=1..N$}{
sample $\bm{\theta}_n \sim \tilde{p}\br{\bm{\theta}}$ \\
sample $\vect{x}_n \sim \prob{\vect{x}\g \bm{\theta}_n}$ \\
}
train $q_{\bm{\phi}}\br{\bm{\theta}\g\vect{x}}$ on $\set{\bm{\theta}_n, \vect{x}_n}$\\
$\hatprob{\bm{\theta}\g \vect{x}=\vect{x}_o} \leftarrow \frac{\prob{\bm{\theta}}}{\tilde{p}\br{\bm{\theta}}}\,q_{\bm{\phi}}\br{\bm{\theta}\g\vect{x}_o}$
\vspace{1px}
\caption{Training of posterior}
\label{alg:train_posterior}
\end{algorithm}
\end{minipage}
\end{tabular}
\vspace*{-0.2cm}
\end{figure}

\subsection{Use of Bayesian neural density estimators}

To make Algorithm~\algref{alg:train_prop_prior} as efficient as possible, the number of simulations per iteration $N$ should be kept small, while at the same time it should provide a sufficient training signal for $q_{\bm{\phi}}$. With a conventional MDN, if $N$ is made too small, there is a danger of overfitting, especially in early iterations, leading to over-confident proposal priors and an unstable procedure. Early stopping could be used to avoid overfitting; however a significant fraction of the $N$ samples would have to be used as a validation set, leading to inefficient use of simulations.

As a better alternative, we developed a Bayesian version of the MDN using Stochastic Variational Inference (SVI) for neural networks \citep{Kingma:2013}. We shall refer to this Bayesian version of the MDN as MDN-SVI\@. An MDN-SVI has two sets of adjustable parameters of the same size, the means $\bm{\phi}_m$ and the log variances $\bm{\phi}_s$. The means correspond to the parameters $\bm{\phi}$ of a conventional MDN\@. During training, Gaussian noise of variance $\exp{\bm{\phi}_s}$ is added to the means independently for each training example $\pair{\bm{\theta}_n}{\vect{x}_n}$.
The Bayesian interpretation of this procedure is that it optimizes a variational Gaussian posterior with a diagonal covariance matrix over parameters $\bm{\phi}$. At prediction time, the noise is switched off and the MDN-SVI behaves like a conventional MDN with $\bm{\phi}=\bm{\phi}_m$.
\ifarxiv
Appendix~\ref{sec:mog_svi}
\else
Section~\supref{suppl-sec:mog_svi} of the supplementary material
\fi
details the implementation and training of MDN-SVI\@.
We found that using an MDN-SVI instead of an MDN improves the robustness and efficiency of Algorithm~\algref{alg:train_prop_prior} because (a)~MDN-SVI is resistant to overfitting, allowing us to use a smaller number of simulations $N$; (b)~no validation set is needed, so all samples can be used for training; and (c)~since overfitting is not an issue, no careful tuning of training time is necessary.

\section{Experiments}
\label{sec:experiments}

We showcase three versions of our approach: (a) learning the posterior with Algorithm~\algref{alg:train_posterior} where $q_{\bm{\phi}}$ is a conventional MDN and the proposal prior $\tilde{p}\br{\bm{\theta}}$ is taken to be the actual prior $\prob{\bm{\theta}}$, which we refer to as \emph{MDN with prior}; (b) training a proposal prior with Algorithm~\algref{alg:train_prop_prior} where $q_{\bm{\phi}}$ is an MDN-SVI, which we refer to as \emph{proposal prior}; and (c) learning the posterior with Algorithm~\algref{alg:train_posterior} where $q_{\bm{\phi}}$ is an MDN-SVI and the proposal prior $\tilde{p}\br{\bm{\theta}}$ is taken to be the one learnt in (b), which we refer to as \emph{MDN with proposal}. All MDNs were trained using Adam \citep{Kingma:2014} with its default parameters.

We compare to three ABC baselines: (a)~rejection ABC \citep{Pritchard:1999}, where parameters are proposed from the prior and are accepted if $\norm{\vect{x}-\vect{x}_o}<\epsilon$; (b)~MCMC-ABC \citep{Marjoram:2003} with a spherical Gaussian proposal, whose variance we manually tuned separately in each case for best performance; and (c)~SMC-ABC \citep{Beaumont:2009}, where the sequence of $\epsilon$'s was exponentially decayed, with a decay rate manually tuned separately in each case for best performance. MCMC-ABC was given the unrealistic advantage of being initialized with a sample from rejection ABC, removing the need for an otherwise necessary burn-in period.
\ifarxiv
Code for reproducing the experiments is provided at \url{https://github.com/gpapamak/epsilon_free_inference}.
\else
Code for reproducing the experiments is provided in the supplementary material and at \url{https://github.com/gpapamak/epsilon_free_inference}.
\fi

\subsection{Mixture of two Gaussians}

The first experiment is a toy problem where the goal is to infer the common mean $\theta$ of a mixture of two 1D Gaussians, given a single datapoint $x_o$. The setup is
\begin{equation}
\prob{\theta} = \uniform{\theta}{\theta_\alpha}{\theta_\beta}
\quad\text{ and }\quad
\prob{x\g\theta} = \alpha\,\gaussian{x}{\theta}{\sigma_1^2} + \br{1-\alpha}\,\gaussian{x}{\theta}{\sigma_2^2},
\end{equation}
where $\theta_\alpha = -10$, $\theta_\beta = 10$, $\alpha=0.5$, $\sigma_1=1$, $\sigma_2=0.1$ and $x_o=0$. The posterior can be calculated analytically, and is proportional to an equal mixture of two Gaussians centred at $x_o$ with variances $\sigma_1^2$ and $\sigma_2^2$, restricted to $\left[\theta_\alpha,\theta_\beta\right]$. This problem is often used in the SMC-ABC literature to illustrate the difficulty of MCMC-ABC in representing long tails. Here we use it to demonstrate the correctness of our approach and its ability to accurately represent non-Gaussian long-tailed posteriors.

\begin{figure}[t]
\def\imwidth{0.327\textwidth}
\centering
\subfloat{
\includegraphics[width=\imwidth]{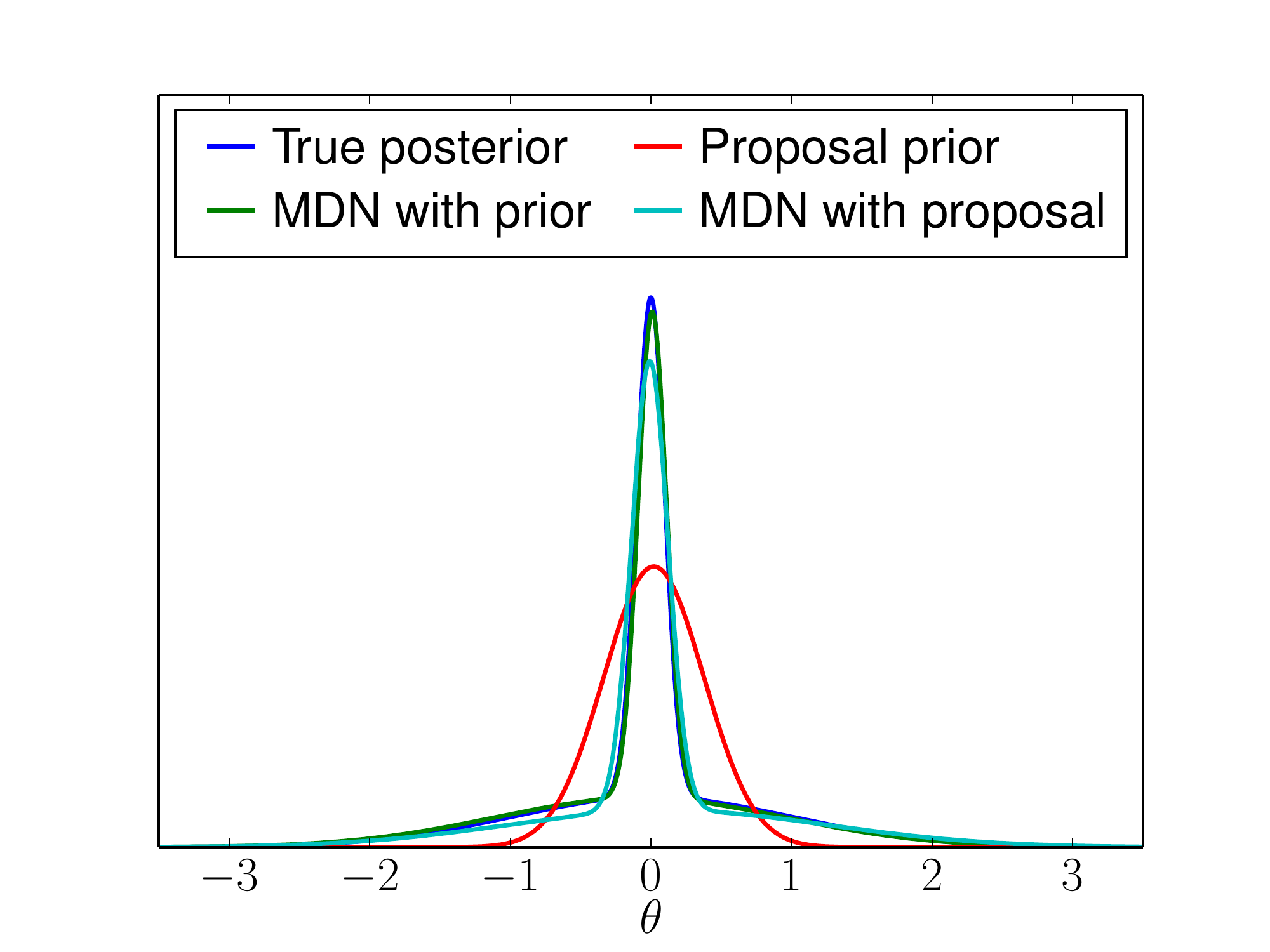}}
\subfloat{
\includegraphics[width=\imwidth]{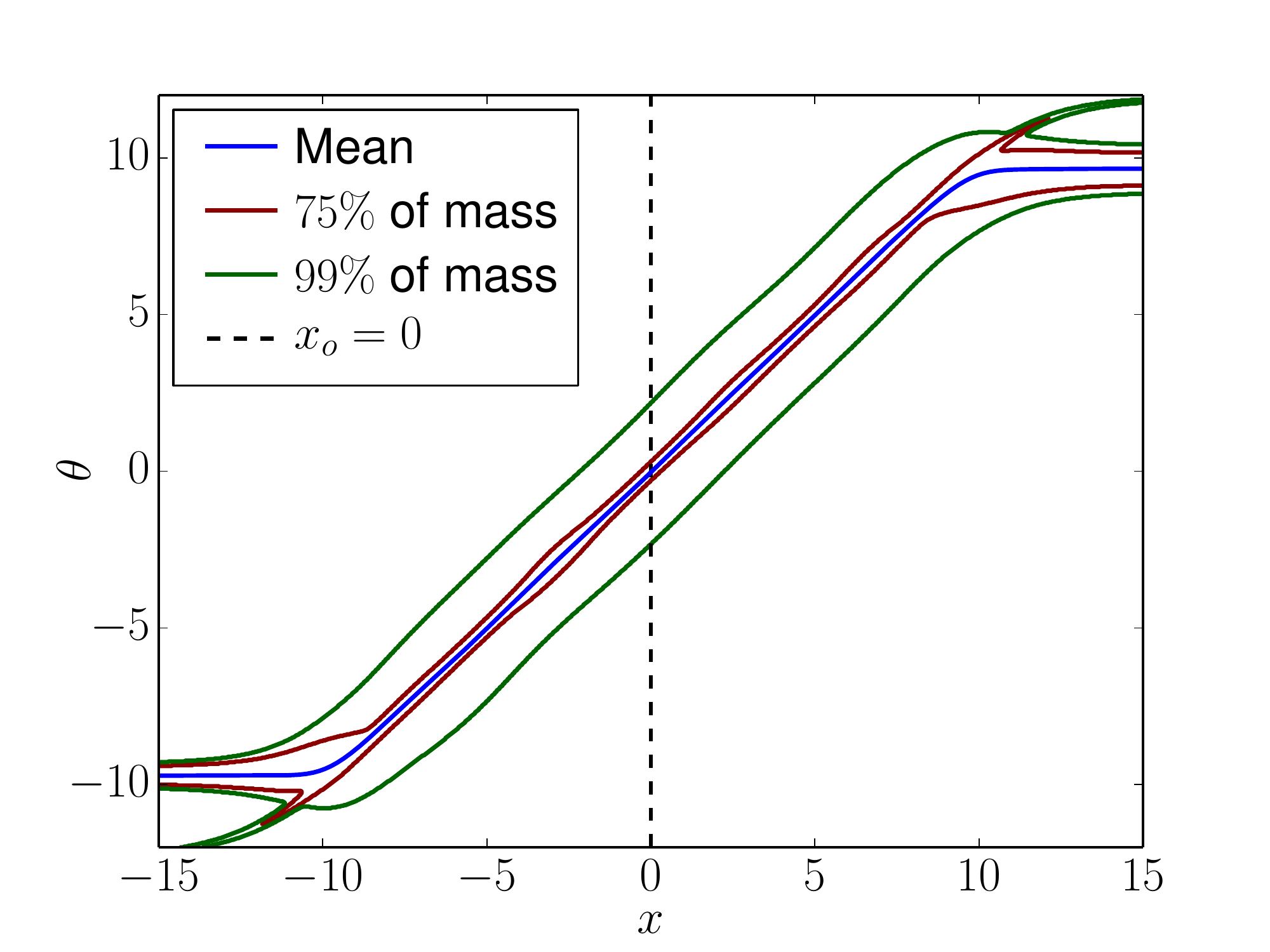}}
\subfloat{
\includegraphics[width=\imwidth]{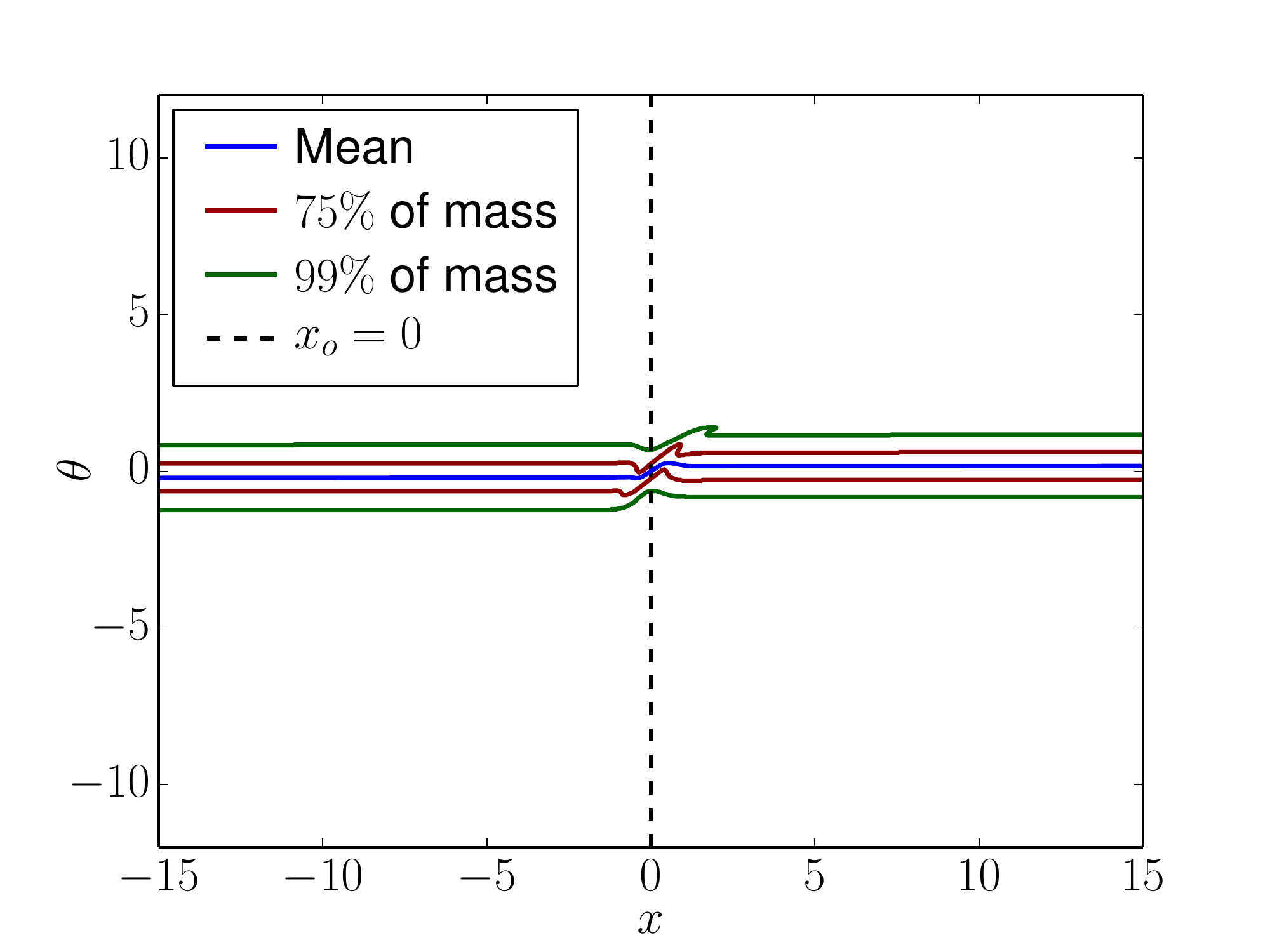}}
\caption{Results on mixture of two Gaussians. \textbf{Left}: approximate posteriors learnt by each strategy for $x_o=0$. \textbf{Middle}: full conditional density $q_{\bm{\phi}}\br{\theta|x}$ leant by the MDN trained with prior. \textbf{Right}: full conditional density $q_{\bm{\phi}}\br{\theta|x}$ learnt by the MDN-SVI trained with proposal prior. Vertical dashed lines show the location of the observation $x_o=0$.}
\label{fig:mog_results}
\end{figure}

Figure~\ref{fig:mog_results} shows the results of neural density estimation using each strategy. All MDNs have one hidden layer with $20$ tanh units and $2$ Gaussian components, except for the proposal prior MDN which has a single component. Both MDN with prior and MDN with proposal learn good parametric approximations to the true posterior, and the proposal prior is a good Gaussian approximation to it.
We used $10$K simulations to train the MDN with prior, whereas the prior proposal took $4$ iterations of $200$ simulations each to train, and the MDN with proposal took $1000$ simulations on top of the previous $800$. The MDN with prior learns the posterior distributions for a large range of possible observations $x$ (middle plot of Figure~\ref{fig:mog_results}), whereas the MDN with proposal gives accurate posterior probabilities only near the value actually observed (right plot of Figure~\ref{fig:mog_results}).

\subsection{Bayesian linear regression}

In Bayesian linear regression, the goal is to infer the parameters $\bm{\theta}$ of a linear map from noisy observations of outputs at known inputs. The setup is
\begin{equation}
\prob{\bm{\theta}} = \gaussian{\bm{\theta}}{\vect{m}}{\mat{S}}
\quad\text{ and }\quad
\prob{\vect{x}\g\bm{\theta}} = \textstyle{\prod_i{\gaussian{x_i}{\bm{\theta}^T\vect{u}_i}{\sigma^2}}},
\end{equation}
where we took $\vect{m}=\vect{0}$, $\mat{S}=\mat{I}$, $\sigma=0.1$, randomly generated inputs $\set{\vect{u}_i}$ from a standard Gaussian, and randomly generated observations $\vect{x}_o$ from the model. In our setup, $\bm{\theta}$ and $\vect{x}$ have $6$ and $10$ dimensions respectively. The posterior is analytically tractable, and is a single Gaussian.

\begin{figure}[t]
\def\imwidth{0.327\textwidth}
\centering
\subfloat{
\includegraphics[width=\imwidth]{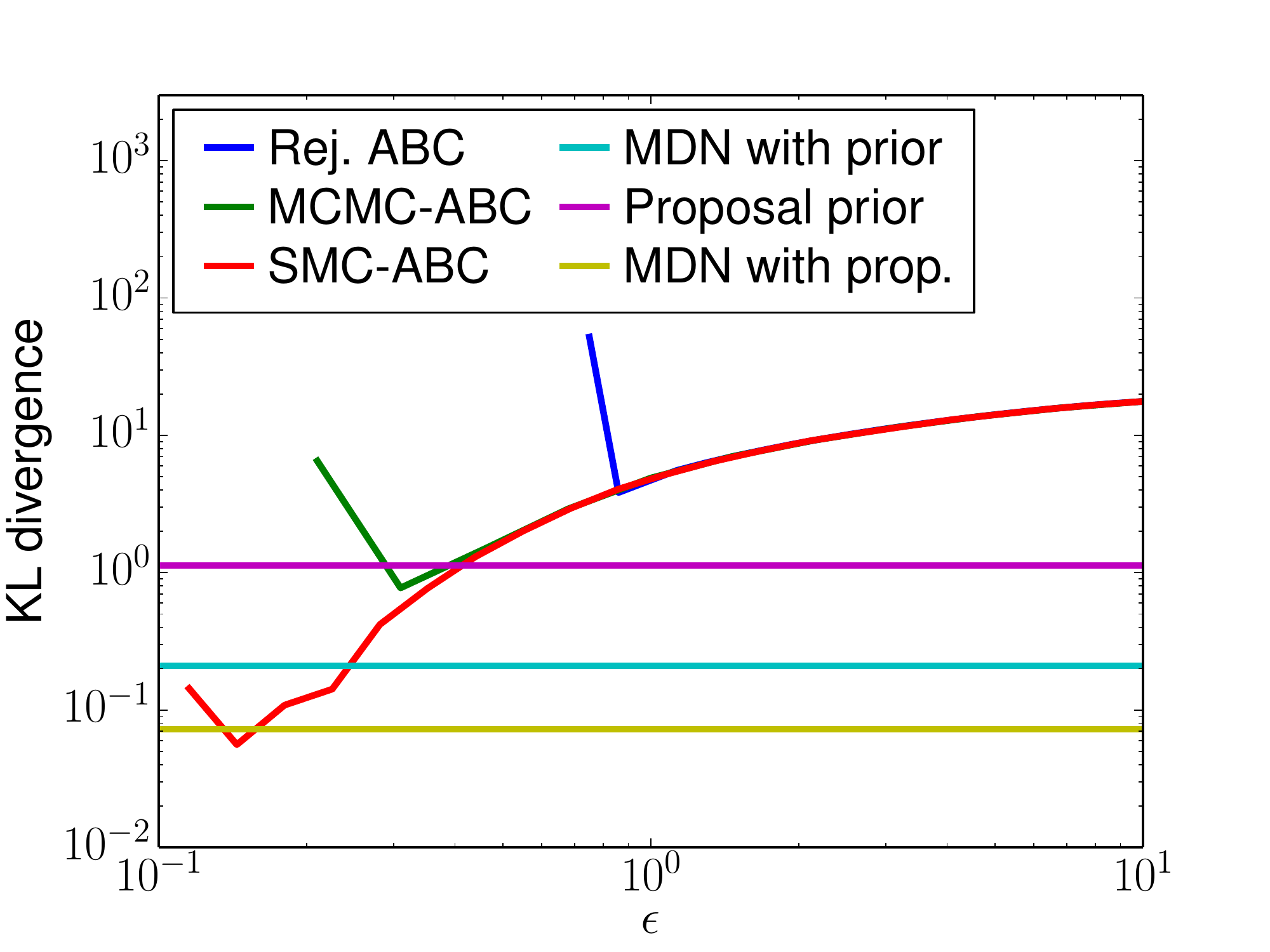}}
\subfloat{
\includegraphics[width=\imwidth]{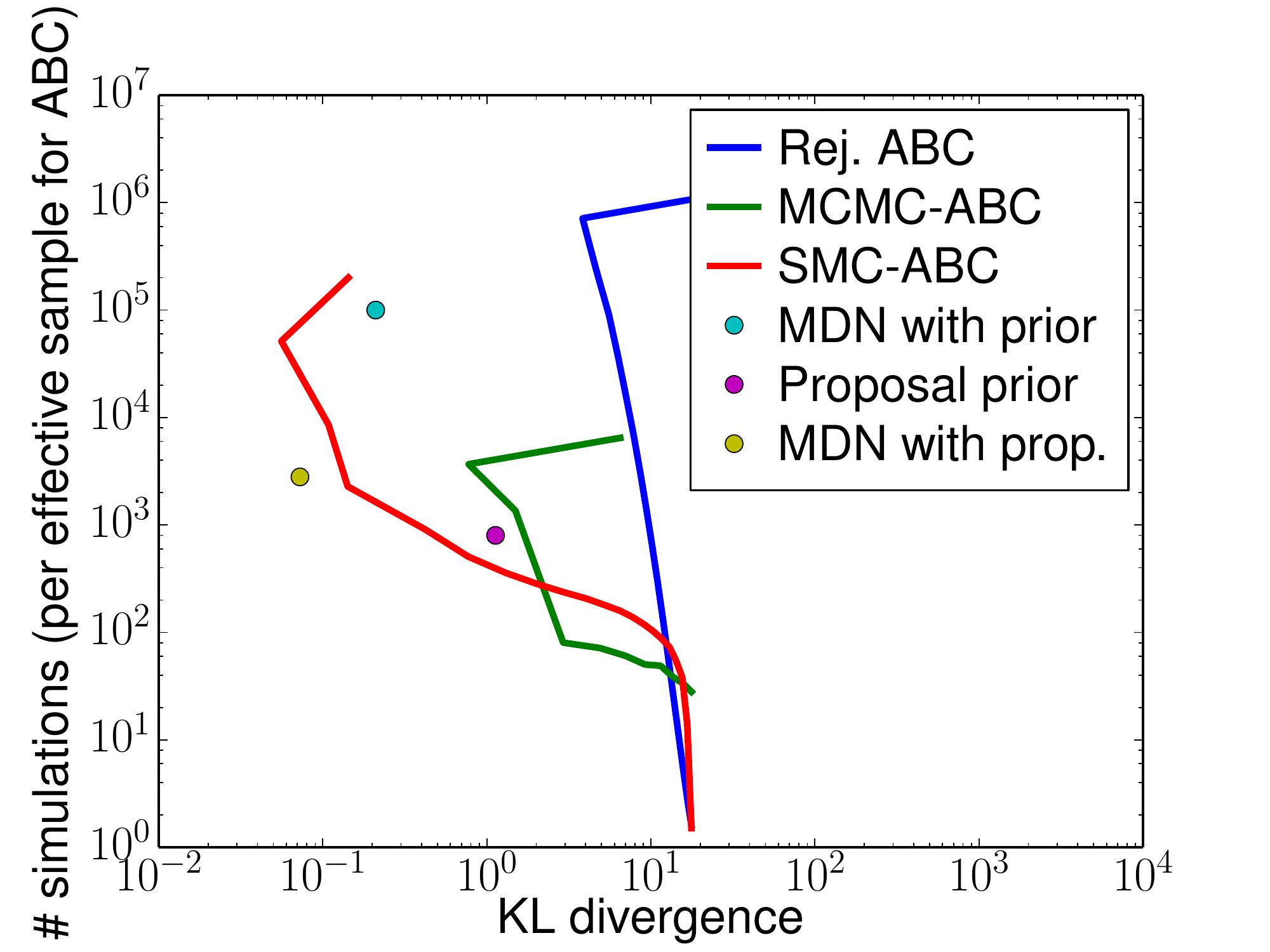}}
\subfloat{
\includegraphics[width=\imwidth]{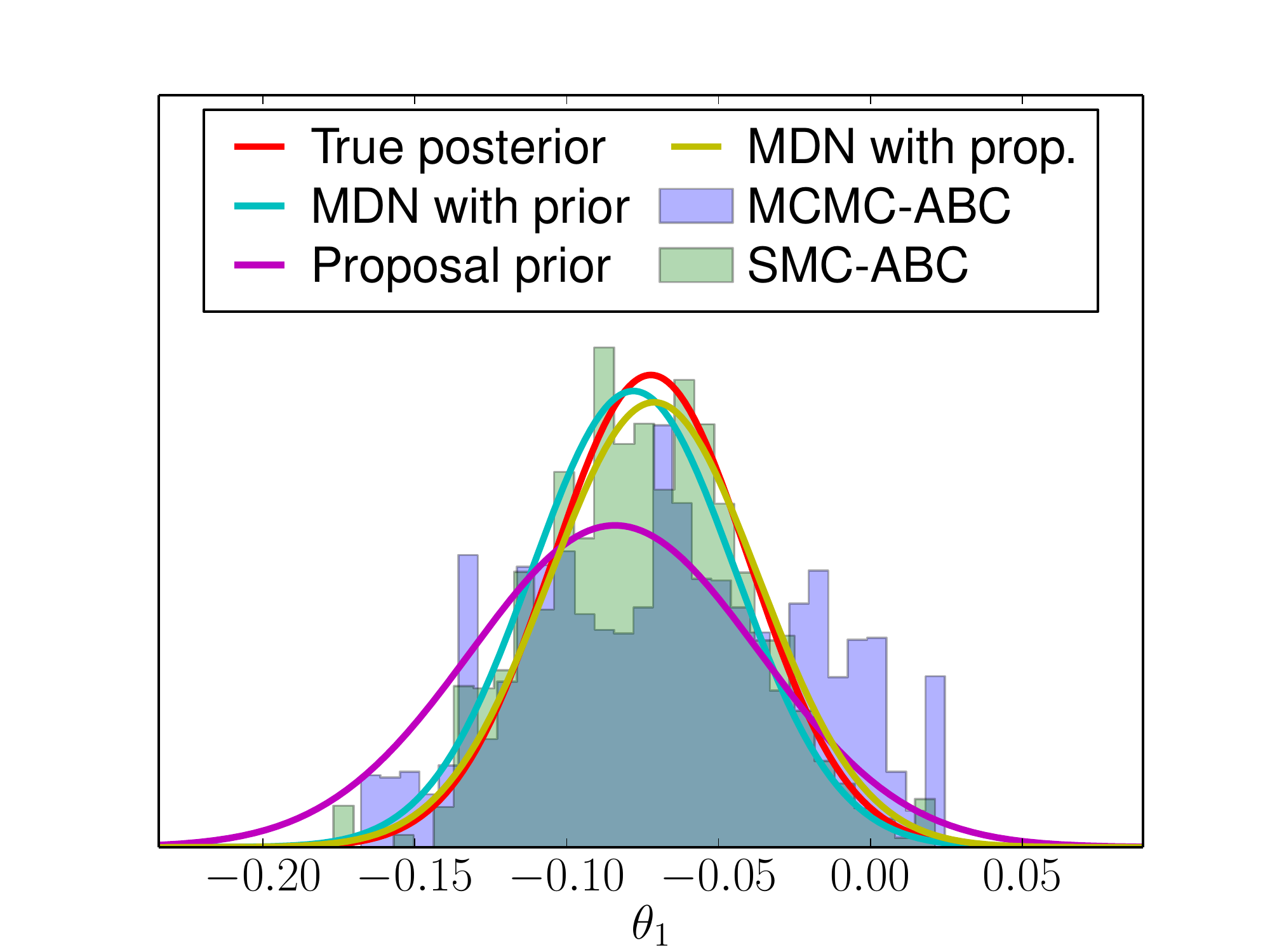}}
\caption{Results on Bayesian linear regression. \textbf{Left}: KL divergence from true posterior to approximation vs $\epsilon$; lower is better. \textbf{Middle}: number of simulations vs KL divergence; lower left is better. Note that number of simulations is total for MDNs, and per effective sample for ABC\@. \textbf{Right}: Posterior marginals for $\theta_1$ as computed by each method. ABC posteriors (represented as histograms) correspond to the setting of $\epsilon$ that minimizes the KL in the left plot.}
\label{fig:blr_results}
\end{figure}

All MDNs have one hidden layer of $50$ tanh units and one Gaussian component. ABC methods were run for a sequence of decreasing $\epsilon$'s, up to their failing points. To measure the approximation quality to the posterior, we analytically calculated the KL divergence from the true posterior to the learnt posterior (which for ABC was taken to be a Gaussian fit to the set of returned posterior samples).
The left of Figure~\ref{fig:blr_results} shows the approximation quality vs $\epsilon$; MDN methods are shown as horizontal lines. As $\epsilon$ is decreased, ABC methods sample from an increasingly better approximation to the true posterior, however they eventually reach their failing point, or take prohibitively long. The best approximations are achieved by MDN with proposal and a very long run of SMC-ABC\@.

The middle of Figure~\ref{fig:blr_results} shows the increase in number of simulations needed to improve approximation quality (as $\epsilon$ decreases). We quote the \emph{total} number of simulations for MDN training, and the number of simulations \emph{per effective sample} for ABC\@.
\ifarxiv
Appendix~\ref{sec:ess_abc}
\else
Section~\supref{suppl-sec:ess_abc} of the supplementary material
\fi
describes how the number of effective samples is calculated. The number of simulations per effective sample should be multiplied by the number of effective samples needed in practice. Moreover, SMC-ABC will not work well with only one particle, so many times the quoted cost will always be needed. Here, MDNs make more efficient use of simulations than Monte Carlo ABC methods. Sequentially fitting a prior proposal was more than ten times cheaper than training with prior samples, and more accurate.

\subsection{Lotka--Volterra predator-prey population model}

The Lotka--Volterra model is a stochastic Markov jump process that describes the continuous time evolution of a population of predators interacting with a population of prey. There are four possible reactions: (a)~a predator being born, (b)~a predator dying, (c)~a prey being born, and (d)~a prey being eaten by a predator.
Positive parameters $\bm{\theta} = \left(\theta_1,\theta_2,\theta_3,\theta_4\right)$ control the rate of each reaction.
Given a set of statistics $\vect{x}_o$ calculated from an observed population time series, the objective is to infer $\bm{\theta}$. We used a flat prior over $\log{\bm{\theta}}$, and calculated a set of $9$ statistics $\vect{x}$. The full setup is detailed in
\ifarxiv
Appendix~\ref{sec:lv_setup}.
\else
Section~\supref{suppl-sec:lv_setup} of the supplementary material.
\fi
The Lotka--Volterra model is commonly used in the ABC literature as a realistic model which can be simulated, but whose likelihood is intractable. One of the properties of Lotka--Volterra is that typical nature-like observations only occur for very specific parameter settings, resulting in narrow, Gaussian-like posteriors that are hard to recover.

The MDN trained with prior has two hidden layers of $50$ tanh units each, whereas the MDN-SVI used to train the proposal prior and the MDN-SVI trained with proposal have one hidden layer of $50$ tanh units. All three have one Gaussian component. We found that using more than one components made no difference to the results; in all cases the MDNs chose to use only one component and switch the rest off, which is consistent with our observation about the near-Gaussianity of the posterior.

\begin{figure}[t]
\def\imwidth{0.327\textwidth}
\centering
\subfloat{
\includegraphics[width=\imwidth]{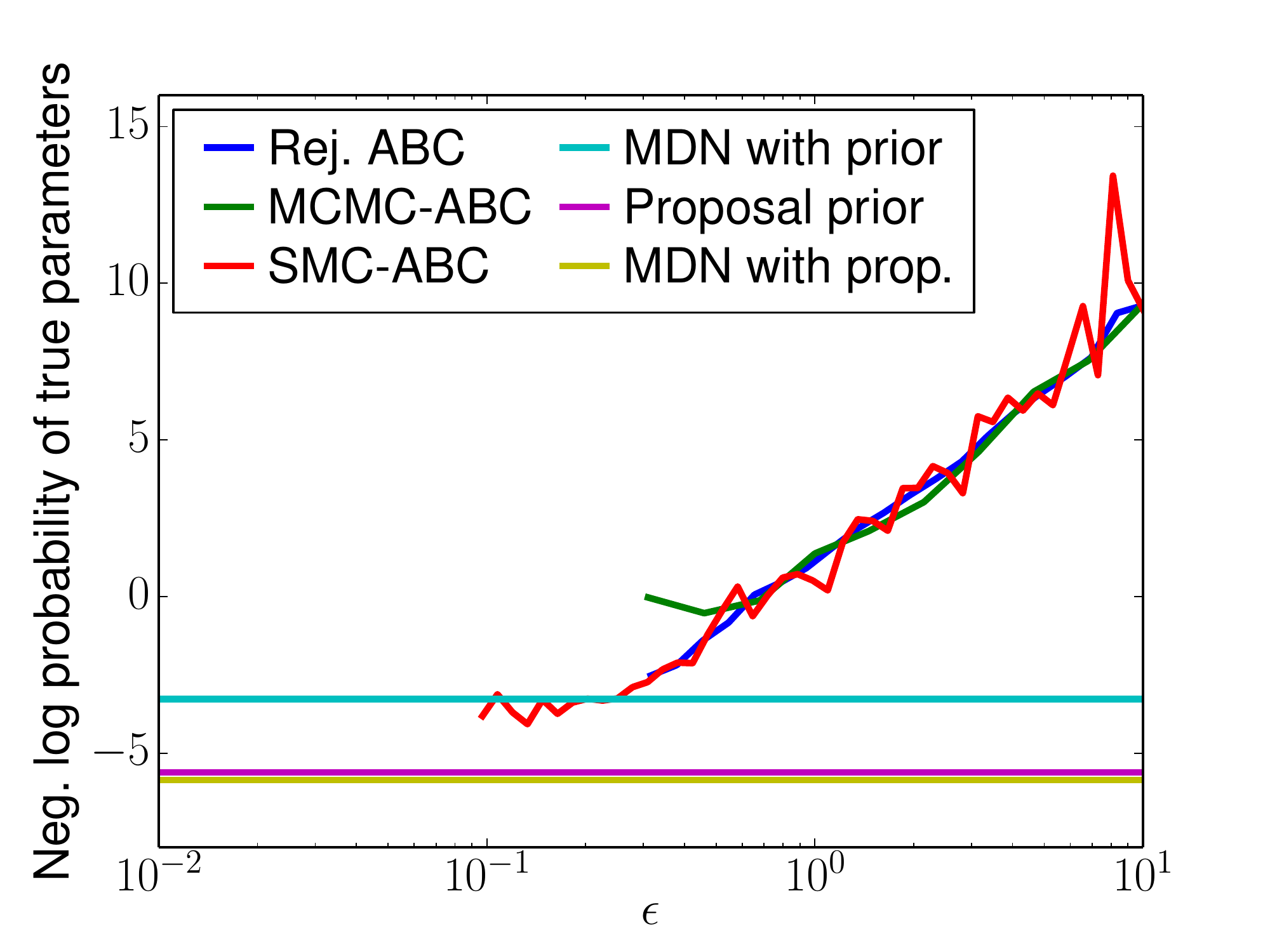}}
\subfloat{
\includegraphics[width=\imwidth]{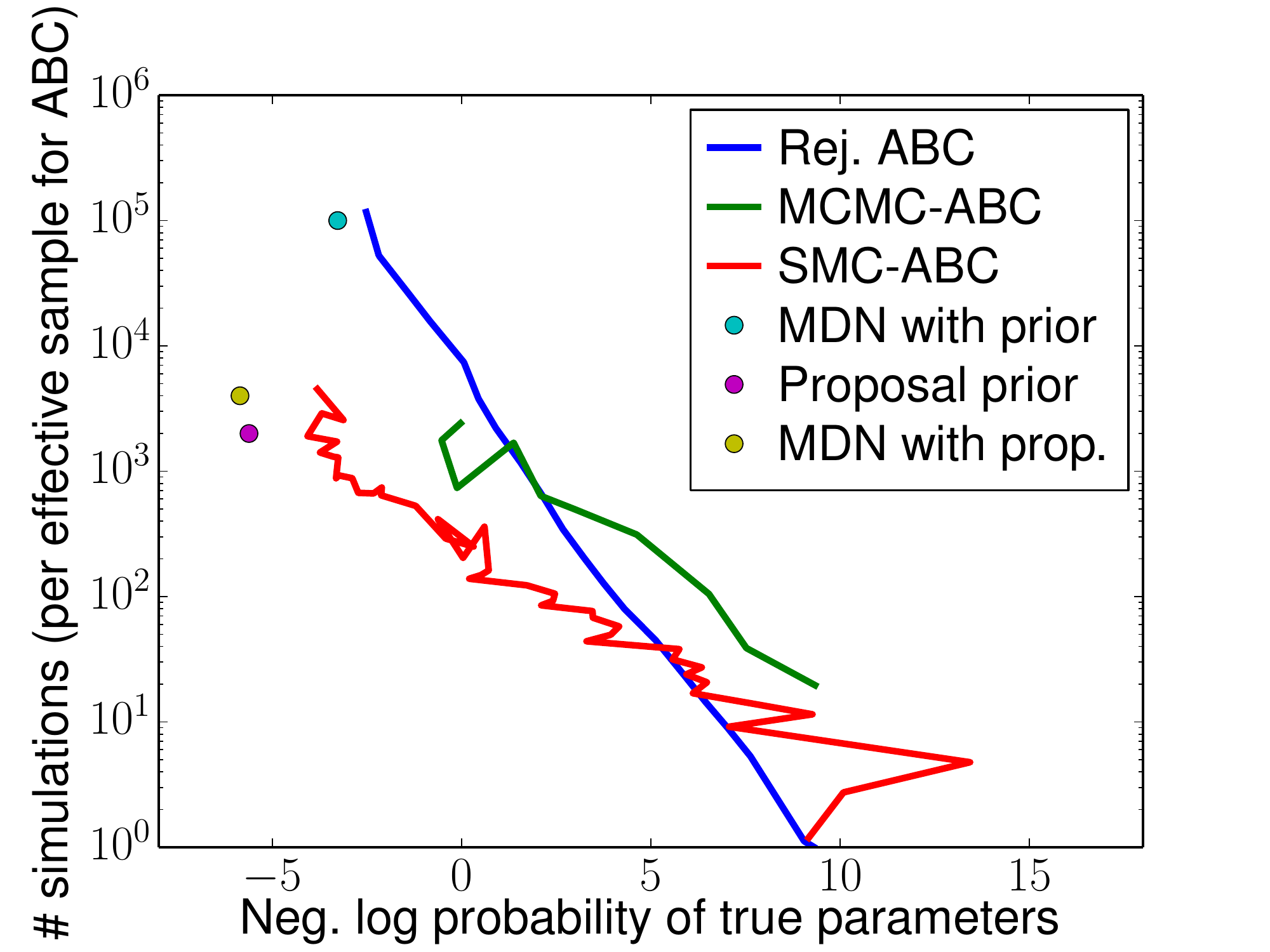}}
\subfloat{
\includegraphics[width=\imwidth]{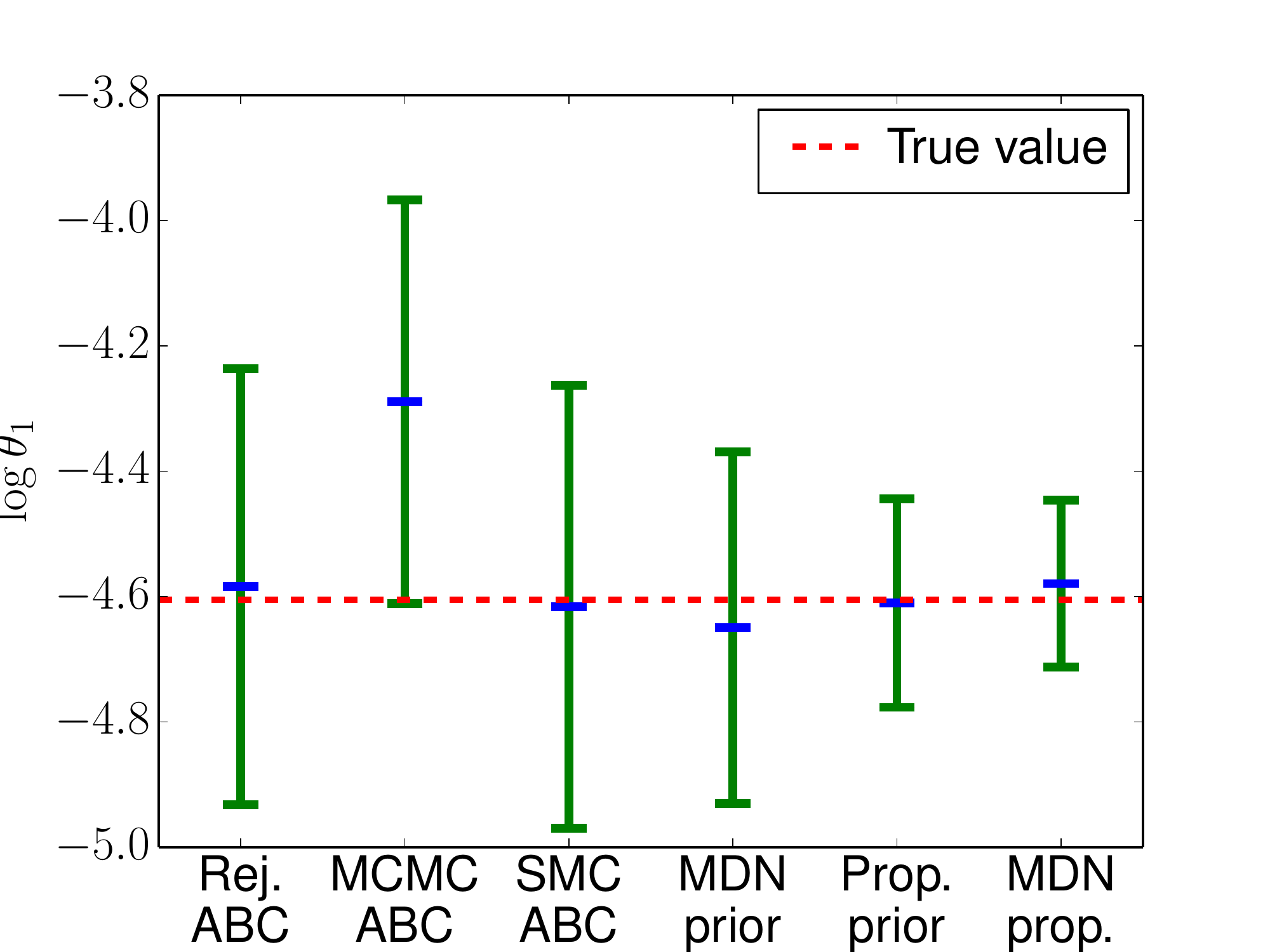}}
\caption{Results on Lotka--Volterra. \textbf{Left}: negative log probability of true parameters vs $\epsilon$; lower is better. \textbf{Middle}: number of simulations vs negative log probability; lower left is better. Note that number of simulations is total for MDNs, but per effective sample for ABC\@. \textbf{Right}: Estimates of $\log{\theta_1}$ with $2$ standard deviations. ABC estimates used many more simulations with the smallest feasible $\epsilon$.\looseness=-1}
\label{fig:lv_results}
\end{figure}

We measure how well each method retrieves the true parameter values that were used to generate $\vect{x}_o$ by calculating their log probability under each learnt posterior; for ABC a Gaussian fit to the posterior samples was used. The left panel of Figure~\ref{fig:lv_results} shows how this log probability varies with~$\epsilon$, demonstrating the superiority of MDN methods over ABC\@. In the middle panel we can see that MDN training with proposal makes efficient use of simulations compared to training with prior and ABC; note that for ABC the number of simulations is only for \emph{one effective sample}. In the right panel, we can see that the estimates returned by MDN methods are more confident around the true parameters compared to ABC, because the MDNs learn the exact posterior rather than an inflated version of it like ABC does (plots for the other three parameters look similar).

We found that when training an MDN with a well-tuned proposal that focuses on the plausible region, an MDN with fewer parameters is needed compared to training with the prior. This is because the MDN trained with proposal needs to learn only the \emph{local} relationship between $\vect{x}$ and $\bm{\theta}$ near $\vect{x}_o$, as opposed to in the entire domain of the prior. Hence, not only are savings achieved in number of simulations, but also training the MDN itself becomes more efficient.

\subsection{M/G/1 queue model}

The M/G/1 queue model describes the processing of a queue of continuously arriving jobs by a single server. In this model, the time the server takes to process each job is independently and uniformly distributed in the interval $\left[\theta_1,\theta_2\right]$. The time interval between arrival of two consecutive jobs is independently and exponentially distributed with rate $\theta_3$. The server observes only the time intervals between departure of two consecutive jobs. Given a set of equally-spaced percentiles $\vect{x}_o$ of inter-departure times, the task is to infer parameters $\bm{\theta} = \left(\theta_1,\theta_2,\theta_3\right)$.
This model is easy to simulate but its likelihood is intractable, and it has often been used as an ABC benchmark \citep{Blum:2010b, Meeds:2015b}. Unlike Lotka--Volterra, data $\vect{x}$ is weakly informative about $\bm{\theta}$, and hence the posterior over $\bm{\theta}$ tends to be broad and non-Gaussian. In our setup, we placed flat independent priors over $\theta_1$, $\theta_2-\theta_1$ and $\theta_3$, and we took $\vect{x}$ to be $5$ equally spaced percentiles, as detailed in
\ifarxiv
Appendix~\ref{sec:mg1_setup}.
\else
Section~\supref{suppl-sec:mg1_setup} of the supplementary material.
\fi

The MDN trained with prior has two hidden layers of $50$ tanh units each, whereas the MDN-SVI used to train the proposal prior and the one trained with proposal have one hidden layer of $50$ tanh units. As observed in the Lotka--Volterra demo, less capacity is required when training with proposal, as the relationship to be learned is local and hence simpler, which saves compute time and gives a more accurate final posterior. All MDNs have $8$ Gaussian components (except the MDN-SVI used to train the proposal prior, which always has one), which, after experimentation, we determined are enough for the MDNs to represent the non-Gaussian nature of the posterior.

\begin{figure}[t]
\def\imwidth{0.327\textwidth}
\centering
\subfloat{
\includegraphics[width=\imwidth]{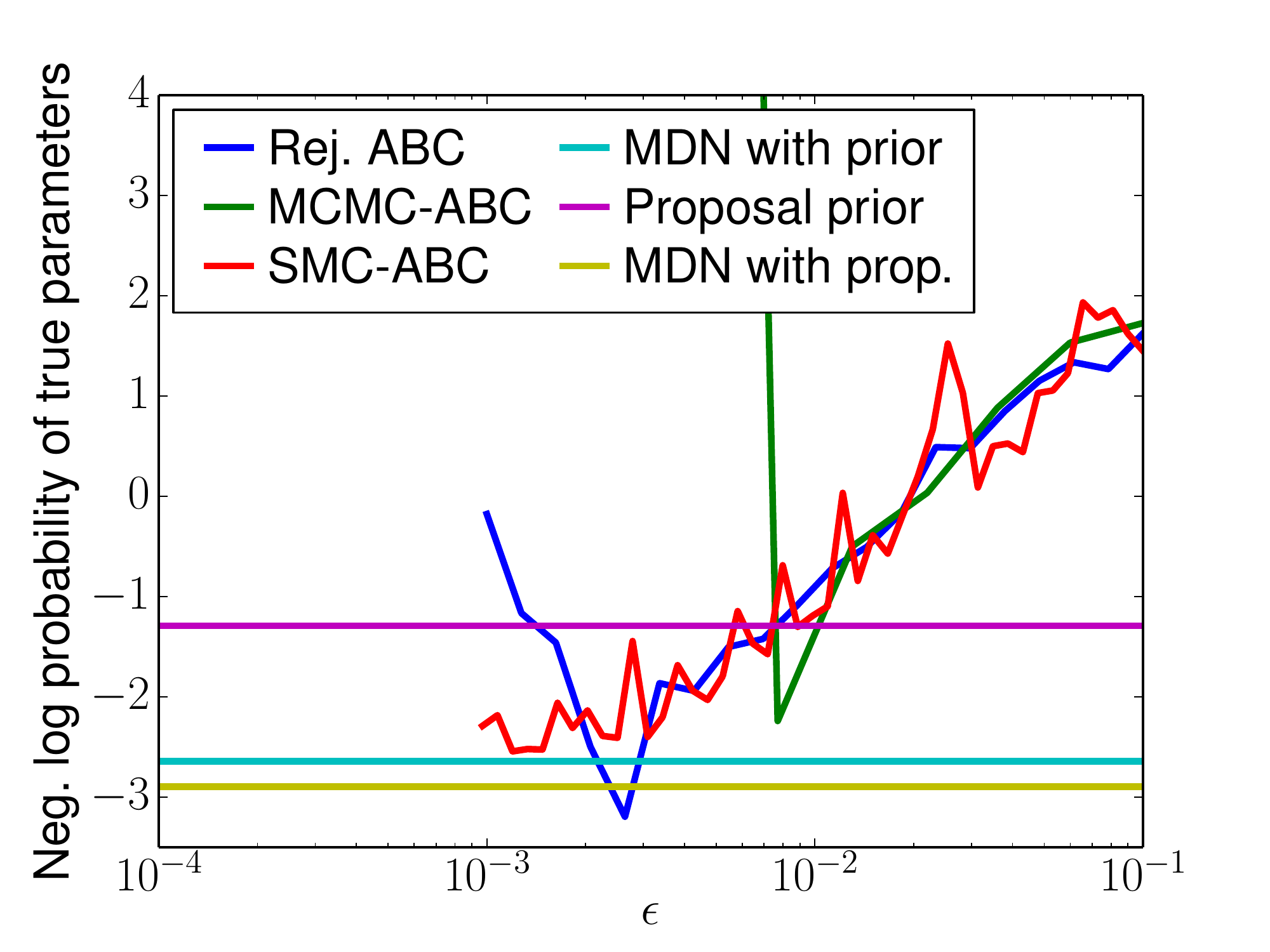}}
\subfloat{
\includegraphics[width=\imwidth]{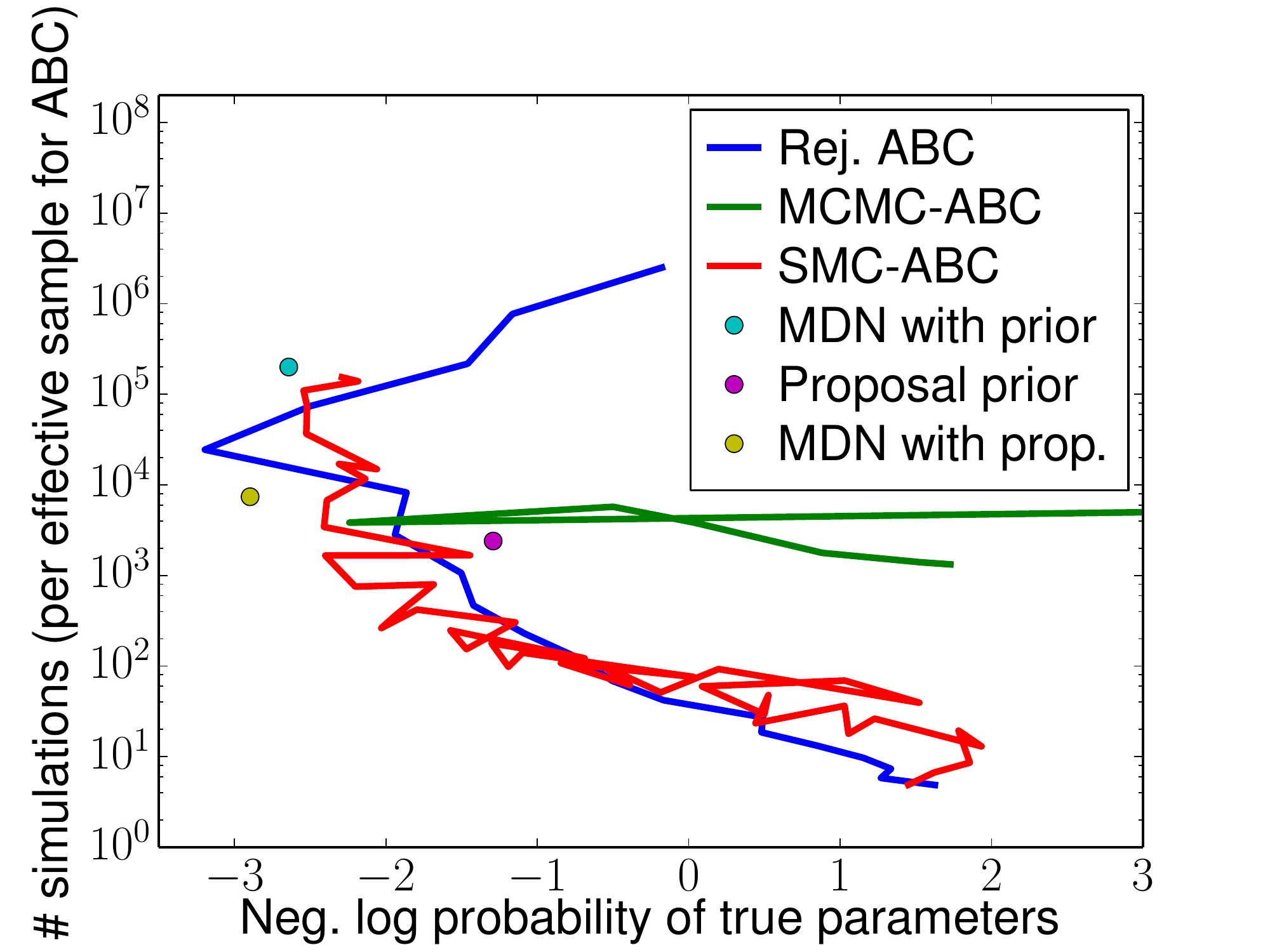}}
\subfloat{
\includegraphics[width=\imwidth]{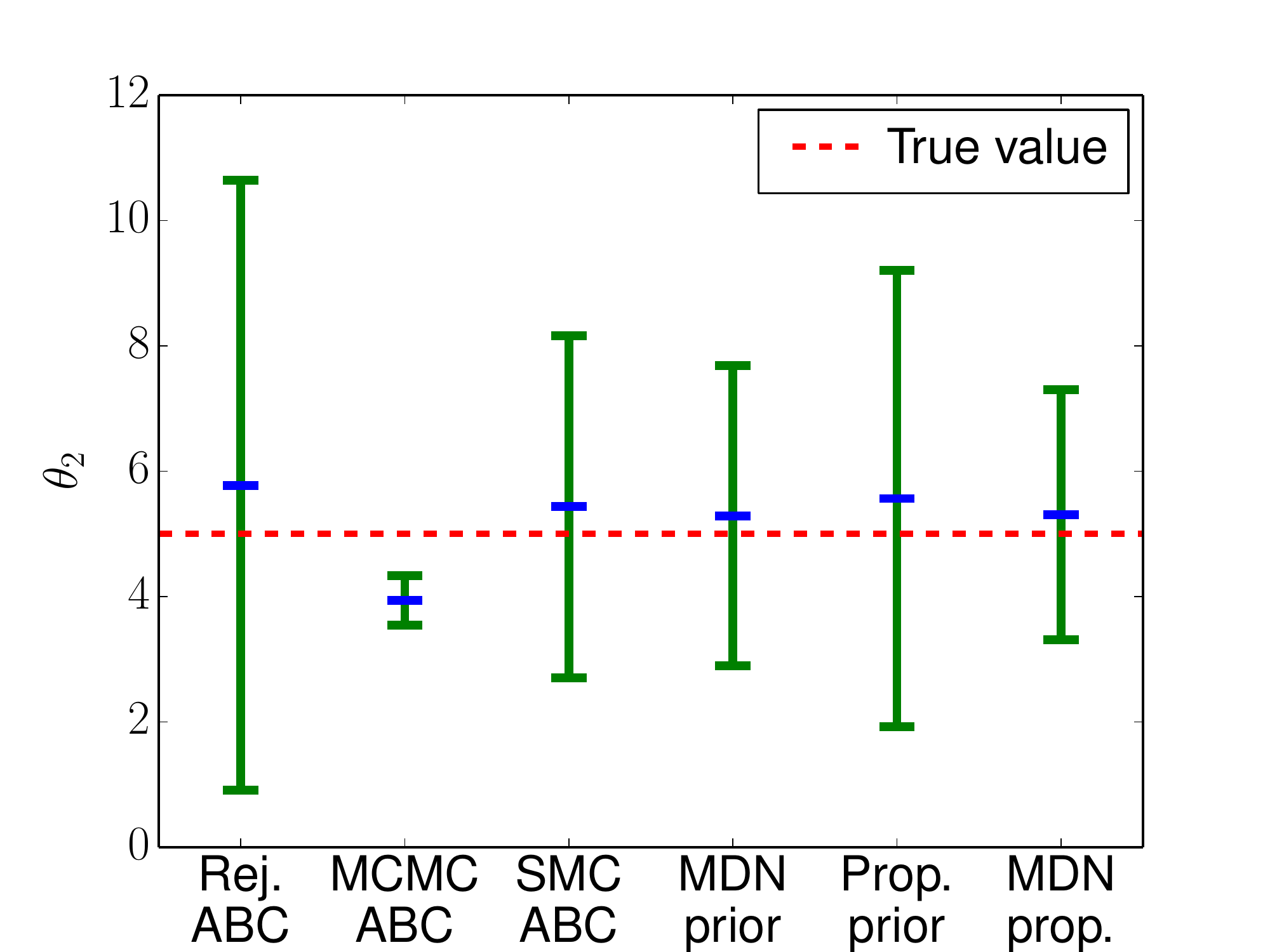}}
\caption{Results on M/G/1. \textbf{Left}: negative log probability of true parameters vs $\epsilon$; lower is better. \textbf{Middle}: number of simulations vs negative log probability; lower left is better. Note that number of simulations is total for MDNs, and per effective sample for ABC\@. \textbf{Right}: Estimates of $\theta_2$ with $2$ standard deviations; ABC estimates correspond to the lowest setting of $\epsilon$ used.}
\label{fig:mg1_results}
\end{figure}

Figure~\ref{fig:mg1_results} reports the log probability of the true parameters under each posterior learnt---for ABC, the log probability was calculated by fitting a mixture of $8$ Gaussians to posterior samples using Expectation-Maximization---and the number of simulations needed to achieve it. As before, MDN methods are more confident compared to ABC around the true parameters, which is due to ABC computing a broader posterior than the true one. MDN methods make more efficient use of simulations, since they use all of them for training and, unlike ABC, do not throw a proportion of them away.

\section{Related work}
\label{sec:related_work}

\textbf{Regression adjustment}\@.
An early parametric approach to ABC is \emph{regression adjustment}, where a parametric regressor is trained on simulation data in order to learn a mapping from $\vect{x}$ to $\bm{\theta}$. The learnt mapping is then used to correct for using a large $\epsilon$, by adjusting the location of posterior samples gathered by e.g.~rejection ABC\@. \citet{Beaumont:2002} used linear regressors, and later  \citet{Blum:2010b} used neural networks with one hidden layer that separately predicted the mean and variance of $\bm{\theta}$. Both can be viewed as rudimentary density estimators and as such they are a predecessor to our work. However, they were not flexible enough to accurately estimate the posterior, and they were only used within some other ABC method to allow for a larger $\epsilon$. In our work, we make conditional density estimation flexible enough to approximate the posterior accurately.

\textbf{Synthetic likelihood}\@.
Another parametric approach is \emph{synthetic likelihood}, where parametric models are used to estimate the likelihood $\prob{\vect{x}\g\bm{\theta}}$. \citet{Wood:2010} used a single Gaussian, and later \citet{Fan:2013} used a mixture Gaussian model. Both of them learnt a separate density model of $\vect{x}$ for each $\bm{\theta}$ by repeatedly simulating the model for fixed $\bm{\theta}$. More recently, \citet{Meeds:2014} used a Gaussian process model to interpolate Gaussian likelihood approximations between different $\bm{\theta}$'s. Compared to learning the posterior, synthetic likelihood has the advantage of not depending on the choice of proposal prior. Its main disadvantage is the need of further approximate inference on top of it in order to obtain the posterior. In our work we directly learn the posterior, eliminating the need for further inference, and we address the problem of correcting for the proposal prior.

\textbf{Efficient Monte Carlo ABC}\@.
Recent work on ABC has focused on reducing the simulation cost of sample-based ABC methods. Hamiltonian ABC \citep{Meeds:2015a} improves upon MCMC-ABC by using stochastically estimated gradients in order to explore the parameter space more efficiently. Optimization Monte Carlo ABC \citep{Meeds:2015b} explicitly optimizes the location of ABC samples, which greatly reduces rejection rate. Bayesian optimization ABC \citep{Gutmann:2015} models $\prob{\norm{\vect{x}-\vect{x}_o}\g\bm{\theta}}$ as a Gaussian process and then uses Bayesian optimization to guide simulations towards the region of small distances $\norm{\vect{x}-\vect{x}_o}$. In our work we show how a significant reduction in simulation cost can also be achieved with parametric methods, which target the posterior directly.

\textbf{Recognition networks}.
Our use of neural density estimators for learning posteriors is reminiscent of recognition networks in machine learning. A recognition network is a neural network that is trained to invert a generative model. The Helmholtz machine \citep{Dayan:1995}, the variational auto-encoder \citep{Kingma:2013} and stochastic backpropagation \citep{Rezende:2014} are examples where a recognition network is trained jointly with the generative network it is designed to invert. Feedforward neural networks have been used to invert black-box generative models \citep{Nair:2008} and binary-valued Bayesian networks \cite{Morris:2001}, and convolutional neural networks have been used to invert a physics engine \citep{Wu:2015}. Our work illustrates the potential of recognition networks in the field of likelihood-free inference, where the generative model is fixed, and inference of its parameters is the goal.

\textbf{Learning proposals}.
Neural density estimators have been employed in learning proposal distributions for importance sampling \citep{Papamakarios:2015} and Sequential Monte Carlo \citep{Gu:2015, Paige:2016}. Although not our focus here, our fit to the posterior could also be used within Monte Carlo inference methods. In this work we see how far we can get purely by fitting a series of conditional density estimators.

\section{Conclusions}
\label{sec:conclusions}

Bayesian conditional density estimation improves likelihood-free inference in three main ways: (a)~it represents the posterior parametrically, as opposed to as a set of samples, allowing for probabilistic evaluations later on in the pipeline; (b)~it targets the exact posterior, rather than an $\epsilon$-approximation of it; and (c)~it makes efficient use of simulations by not rejecting samples, by interpolating between samples, and by gradually focusing on the plausible parameter region. Our belief is that neural density estimation is a tool with great potential in likelihood-free inference, and our hope is that this work helps in establishing its usefulness in the field.

\ifanonymize
\else
\subsubsection*{Acknowledgments}

\vspace*{-0.05in}
We thank Amos Storkey for useful comments.
George Papamakarios is supported by the Centre for Doctoral Training in Data Science, funded by EPSRC (grant EP/L016427/1) and the University of Edinburgh, and by Microsoft Research through its PhD Scholarship Programme.
\vspace*{0.05in}
\fi

\renewcommand{\theHsection}{\Alph{section}}
\appendix

\ifarxiv
\section{Proof of Proposition~\ref{proposition}}
\else
\section{Proof of Proposition~\supref{main-proposition}}
\fi
\label{sec:proof}

Maximizing $\prod_n{q_{\bm{\phi}}\br{\bm{\theta}_n\g \vect{x}_n}}$ w.r.t.~$\bm{\phi}$ is equivalent to maximizing the average log probability
\begin{equation}
\frac{1}{N}\sum_n{\log{q_{\bm{\phi}}\br{\bm{\theta}_n\g \vect{x}_n}}}.
\end{equation}
Since $\pair{\bm{\theta}_n}{\vect{x}_n} \sim \tilde{p}\br{\bm{\theta}}\,\prob{\vect{x}\g \bm{\theta}}$, due to the strong law of large numbers, as $N\rightarrow\infty$ the average log probability converges almost surely to the following expectation
\begin{equation}
\frac{1}{N}\sum_n{\log{q_{\bm{\phi}}\br{\bm{\theta}_n\g \vect{x}_n}}}
\xrightarrow{\mathrm{a.s.}}
\avg{\log{q_{\bm{\phi}}\br{\bm{\theta}\g \vect{x}}}}{\tilde{p}\br{\bm{\theta}}\,\prob{\vect{x}\g \bm{\theta}}}.
\end{equation}
Let $\tilde{p}\br{\vect{x}}$ be a distribution over $\vect{x}$. Maximizing the above expectation w.r.t.~$\bm{\phi}$ is equivalent to minimizing
\begin{equation}
\kl{\tilde{p}\br{\bm{\theta}}\,\prob{\vect{x}\g \bm{\theta}}}{\tilde{p}\br{\vect{x}}\,q_{\bm{\phi}}\br{\bm{\theta}\g \vect{x}}} = 
-\avg{\log{q_{\bm{\phi}}\br{\bm{\theta}\g \vect{x}}}}{\tilde{p}\br{\bm{\theta}}\,\prob{\vect{x}\g \bm{\theta}}} + \mathrm{const}.
\end{equation}
The above KL divergence is minimized (and becomes $0$) if and only if
\begin{equation}
\tilde{p}\br{\bm{\theta}}\,\prob{\vect{x}\g \bm{\theta}} = \tilde{p}\br{\vect{x}}\,q_{\bm{\phi}}\br{\bm{\theta}\g \vect{x}}
\end{equation}
almost everywhere. It is easy to see that this can only happen for $\tilde{p}\br{\vect{x}} = \int{\tilde{p}\br{\bm{\theta}}\,\prob{\vect{x}\g \bm{\theta}}\,\mathrm{d}\bm{\theta}}$, since
\begin{equation}
\tilde{p}\br{\bm{\theta}}\,\prob{\vect{x}\g \bm{\theta}} = \tilde{p}\br{\vect{x}}\,q_{\bm{\phi}}\br{\bm{\theta}\g \vect{x}} \;\Rightarrow\;
\int{\tilde{p}\br{\bm{\theta}}\,\prob{\vect{x}\g \bm{\theta}}\,\mathrm{d}\bm{\theta}} = \tilde{p}\br{\vect{x}}\int{q_{\bm{\phi}}\br{\bm{\theta}\g \vect{x}}\,\mathrm{d}\bm{\theta}} = \tilde{p}\br{\vect{x}}.
\end{equation}
Thus, taking $\tilde{p}\br{\vect{x}}$ as above, and assuming a setting of $\bm{\phi}$ that makes the KL equal to $0$ exists, the KL is minimized if and only if we have almost everywhere that
\begin{equation}
q_{\bm{\phi}}\br{\bm{\theta}\g \vect{x}} = 
\frac{\tilde{p}\br{\bm{\theta}}}{\tilde{p}\br{\vect{x}}}\,\prob{\vect{x}\g \bm{\theta}}
= \frac{\tilde{p}\br{\bm{\theta}}}{\tilde{p}\br{\vect{x}}}\,\frac{\prob{\bm{\theta}\g\vect{x}}\,\prob{\vect{x}}}{\prob{\bm{\theta}}}
\propto \frac{\tilde{p}\br{\bm{\theta}}}{\prob{\bm{\theta}}}\,\prob{\bm{\theta}\g \vect{x}}.
\end{equation}
A corollary of the above is that
\begin{equation}
q_{\bm{\phi}}\br{\bm{\theta}\g \vect{x}} = 
\frac{\tilde{p}\br{\bm{\theta}}\,\prob{\vect{x}\g \bm{\theta}}}{\int{\tilde{p}\br{\bm{\theta}}\,\prob{\vect{x}\g \bm{\theta}}\,\mathrm{d}\bm{\theta}}},
\end{equation}
in other words, $q_{\bm{\phi}}\br{\bm{\theta}\g \vect{x}}$ becomes what the posterior would be if the prior were $\tilde{p}\br{\bm{\theta}}$.

\section{Parameterization and training of Mixture Density Networks}
\label{sec:mdn_parameterization}

A Mixture Density Network (MDN) \citep{Bishop:1994} is a conditional density estimator $q_{\bm{\phi}}\br{\bm{\theta}\g\vect{x}}$, which takes the form of a mixture of $K$ Gaussian components, as follows
\begin{equation}
q_{\bm{\phi}}\br{\bm{\theta}\g\vect{x}} = \sum_k{\alpha_k\,\gaussian{\bm{\theta}}{\vect{m}_k}{\mat{S}_k}}.
\end{equation}
The mixing coefficients $\bm{\alpha}=\br{\alpha_1,\ldots,\alpha_K}$, means $\set{\vect{m}_k}$ and covariance matrices $\set{\mat{S}_k}$ are computed by a feedforward neural network $f_{\mat{W},\vect{b}}\br{\vect{x}}$, which has input $\vect{x}$, weights $\mat{W}$ and biases $\vect{b}$. In particular, let the output of the neural network be
\begin{equation}
\vect{y} = f_{\mat{W},\vect{b}}\br{\vect{x}}.
\end{equation}
Then, the mixing coefficients are given by
\begin{equation}
\bm{\alpha} = \mathrm{softmax}\br{\mat{W}_{\bm{\alpha}}\vect{y} + \vect{b}_{\bm{\alpha}}}.
\end{equation}
The softmax ensures that the mixing coefficients are strictly positive and sum to one.
Similarly, the means are given by
\begin{equation}
\vect{m}_k = \mat{W}_{\vect{m}_k}\vect{y} + \vect{b}_{\vect{m}_k}.
\end{equation}
As for the covariance matrices, we need to ensure that they are symmetric and positive definite. For this reason, instead of parameterizing the covariance matrices directly, we parameterize the Cholesky factorization of their inverses. That is, we rewrite
\begin{equation}
\mat{S}_k^{-1} = \mat{U}_k^T \mat{U}_k,
\end{equation}
where $\mat{U}_k$ is parameterized to be an upper triangular matrix with strictly positive elements in the diagonal, as follows
\begin{align}
\mathrm{diag}\br{\mat{U}_k} &= \exp\br{\mat{W}_{\mathrm{diag}\br{\mat{U}_k}}\vect{y} + \vect{b}_{\mathrm{diag}\br{\mat{U}_k}}} \\
\mathrm{utri}\br{\mat{U}_k} &= \mat{W}_{\mathrm{utri}\br{\mat{U}_k}}\vect{y} + \vect{b}_{\mathrm{utri}\br{\mat{U}_k}} \\
\mathrm{ltri}\br{\mat{U}_k} &= \mat{0}.
\end{align}
In the above, $\mathrm{diag}\br{\cdot}$ picks out the diagonal elements, whereas $\mathrm{utri}\br{\cdot}$ and $\mathrm{ltri}\br{\cdot}$ pick out the elements above and below the diagonal respectively. We chose to parameterize the factorization of $\mat{S}_k^{-1}$ rather than that of $\mat{S}_k$, since it is the inverse covariance that directly appears in the calculation of $\gaussian{\bm{\theta}}{\vect{m}_k}{\mat{S}_k}$. Apart from ensuring the symmetry and positive definiteness of $\mat{S}_k$, the above parameterization also allows for efficiently calculating the log determinant of $\mat{S}_k$ as follows
\begin{equation}
-\frac{1}{2}\log{\det\br{\mat{S}_k}} = \mathrm{sum}\br{\mat{W}_{\mathrm{diag}\br{\mat{U}_k}}\vect{y} + \vect{b}_{\mathrm{diag}\br{\mat{U}_k}}}.
\end{equation}
The above parameterization of the covariance matrix was introduced by \citet{Williams:1996} for learning conditional Gaussians.

Given a set of training data $\set{\bm{\theta}_n,\vect{x}_n}$, training the MDN with maximum likelihood amounts to maximizing the average log probability
\begin{equation}
\frac{1}{N}\sum_n{\log{q_{\bm{\phi}}\br{\bm{\theta}_n\g\vect{x}_n}}}
\end{equation}
with respect to the MDN parameters
\begin{equation}
\bm{\phi} = \br{\mat{W},\vect{b},\mat{W}_{\bm{\alpha}},\vect{b}_{\bm{\alpha}},\set{\mat{W}_{\vect{m}_k},\vect{b}_{\vect{m}_k},\mat{W}_{\mathrm{diag}\br{\mat{U}_k}},\vect{b}_{\mathrm{diag}\br{\mat{U}_k}},\mat{W}_{\mathrm{utri}\br{\mat{U}_k}},\vect{b}_{\mathrm{utri}\br{\mat{U}_k}}}}.
\end{equation}
Because the reparameterization $\bm{\phi}$ described above is unconstrained, any off-the-shelf gradient-based stochastic optimizer can be used. Gradients of the average log probability can be easily computed with backpropagation. In our experiments, we implemented MDNs using Theano~\citep{Theano:2016}, which automatically backpropagates gradients, and we maximized the average log likelihood using Adam \citep{Kingma:2014}, which is currently the state of the art in minibatch-based stochastic optimization.

\section{Analytical calculation of parameter posterior}
\label{sec:mog_calc_posterior}

According to
\ifarxiv
Proposition~\ref{proposition},
\else
Proposition~\supref{main-proposition},
\fi
after training $q_{\bm{\phi}}\br{\bm{\theta}\g\vect{x}}$, the posterior at $\vect{x}=\vect{x}_o$ is approximated by
\begin{equation}
\hatprob{\bm{\theta}\g \vect{x}=\vect{x}_o} \propto\frac{\prob{\bm{\theta}}}{\tilde{p}\br{\bm{\theta}}}\,q_{\bm{\phi}}\br{\bm{\theta}\g\vect{x}_o}.
\end{equation}
Typically, the prior $\prob{\bm{\theta}}$ is a simple distribution like a uniform or a Gaussian. Here we will consider the uniform case, while the Gaussian case is treated analogously. Let $\prob{\bm{\theta}}$ be uniform everywhere (improper). Then the posterior estimate becomes
\begin{equation}
\hatprob{\bm{\theta}\g \vect{x}=\vect{x}_o} \propto \frac{q_{\bm{\phi}}\br{\bm{\theta}\g\vect{x}_o}}{\tilde{p}\br{\bm{\theta}}}.
\end{equation}
In practice, we also used this estimate for uniform priors with broad but finite support.
Since $q_{\bm{\phi}}\br{\bm{\theta}\g\vect{x}_o}$ is a mixture of $K$ Gaussians and $\tilde{p}\br{\bm{\theta}}$ is a single Gaussian, that is
\begin{equation}
q_{\bm{\phi}}\br{\bm{\theta}\g\vect{x}} = \sum_k{\alpha_k\,\gaussian{\bm{\theta}}{\vect{m}_k}{\mat{S}_k}}
\quad\text{ and }\quad
\tilde{p}\br{\bm{\theta}} = \gaussian{\bm{\theta}}{\vect{m}_0}{\mat{S}_0},
\end{equation}
their ratio can be calculated and normalized analytically. In particular, after some algebra it can be shown that the posterior estimate $\hatprob{\bm{\theta}\g \vect{x}=\vect{x}_o}$ is also a mixture of $K$ Gaussians
\begin{equation}
\hatprob{\bm{\theta}\g \vect{x}=\vect{x}_o} = \sum_k{\alpha_k'\,\gaussian{\bm{\theta}}{\vect{m}_k'}{\mat{S}_k'}},
\end{equation}
whose parameters are
\begin{align}
\mat{S}_k' &= \br{\mat{S}_k^{-1} - \mat{S}_0^{-1}}^{-1} \\
\vect{m}_k' &= \mat{S}_k'\br{\mat{S}_k^{-1}\vect{m}_k - \mat{S}_0^{-1}\vect{m}_0} \\
\alpha_k' &= \frac{\alpha_k\exp{\br{-\frac{1}{2}c_k}}}{\sum_{k'}{\alpha_{k'}\exp{\br{-\frac{1}{2}c_{k'}}}}},
\end{align}
where quantities $\set{c_k}$ are given by
\begin{equation}
c_k = \log{\det{\mat{S}_k}} - \log{\det{\mat{S}_0}} - \log{\det{\mat{S}_k'}} + \vect{m}_k^T\mat{S}_k^{-1}\vect{m}_k - \vect{m}_0^T\mat{S}_0^{-1}\vect{m}_0 - \vect{m}_k'^T\mat{S}_k'^{-1}\vect{m}_k'.
\end{equation}
For the above mixture to be well defined, the covariance matrices $\set{\mat{S}_k'}$ must be positive definite. This will not be the case if the proposal prior $\tilde{p}\br{\bm{\theta}}$ is narrower than some component of $q_{\bm{\phi}}\br{\bm{\theta}\g\vect{x}_o}$ along some dimension. However, in both
\ifarxiv
Algorithms~\algref{alg:train_prop_prior} and \algref{alg:train_posterior},
\else
Algorithms~\supref{main-alg:train_prop_prior} and \supref{main-alg:train_posterior},
\fi
$q_{\bm{\phi}}\br{\bm{\theta}\g\vect{x}_o}$ is trained on parameters sampled from $\tilde{p}\br{\bm{\theta}}$, hence, if trained properly, it tends to be narrower than $\tilde{p}\br{\bm{\theta}}$. Our experience with
\ifarxiv
Algorithms~\algref{alg:train_prop_prior} and \algref{alg:train_posterior}
\else
Algorithms~\supref{main-alg:train_prop_prior} and \supref{main-alg:train_posterior}
\fi
is that $\set{\mat{S}_k'}$ not being positive definite rarely happens, whereas it happening is an indication that the algorithm's parameters have not been set up properly.

\section{Stochastic Variational Inference for Mixture Density Networks}
\label{sec:mog_svi}

In this section we describe our adaptation of Stochastic Variational Inference (SVI) for neural networks \citep{Kingma:2013}, in order to develop a Bayesian version of MDN\@. The first step is to express beliefs about the MDN parameters $\bm{\phi}$ as independent Gaussian random variables with means $\bm{\phi}_m$ and log variances $\bm{\phi}_s$. Under this interpretation we can rewrite the parameters as
\begin{equation}
\bm{\phi} = \bm{\phi}_m + \exp{\br{\frac{1}{2}\bm{\phi}_s}}\odot\vect{u},
\end{equation}
where the symbol $\odot$ denotes elementwise multiplication and $\vect{u}$ is an unknown vector drawn from a standard normal,
\begin{equation}
\vect{u} \sim \gaussian{\vect{u}}{\vect{0}}{\mat{I}}.
\end{equation}
The above parameterization induces the following variational distribution over $\bm{\phi}$
\begin{equation}
q\br{\bm{\phi}} = \gaussian{\bm{\phi}}{\bm{\phi}_m}{\mathrm{diag}\br{\exp{\bm{\phi}_s}}},
\end{equation}
where $\mathrm{diag}\br{\exp{\bm{\phi}_s}}$ denotes a diagonal covariance matrix whose diagonal is the vector $\exp{\bm{\phi}_s}$.
Moreover, we place the following Bayesian prior over $\bm{\phi}$
\begin{equation}
\prob{\bm{\phi}} = \gaussian{\bm{\phi}}{\vect{0}}{\lambda^{-1}\vect{I}}.
\end{equation}
Under this prior, before seeing any data we set the parameter means $\bm{\phi}_m$ all to zero, and the parameter log variances $\bm{\phi}_s$ all equal to $\log{\lambda^{-1}}$. In our experiments, we used a default value of $\lambda = 0.01$.

Given training data $\set{\bm{\theta}_n, \vect{x}_n}$, the objective of SVI is to optimize $\bm{\phi}_m$ and $\bm{\phi}_s$ so as to make the variational distribution $q\br{\bm{\phi}}$ be as close as possible (in KL) to the true Bayesian posterior over $\bm{\phi}$. This objective is equivalent to maximizing a variational lower bound,
\begin{equation}
\frac{1}{N}\sum_n{\avg{\log{q_{\bm{\phi}}\br{\bm{\theta}_n\g\vect{x}_n}}}{\gaussian{\vect{u}}{\vect{0}}{\mat{I}}}}
- \frac{1}{N}\kl{q\br{\bm{\phi}}}{\prob{\bm{\phi}}},
\end{equation}
with respect to $\bm{\phi}_m$ and $\bm{\phi}_s$.
The expectations in the first term of the above can be stochastically approximated by randomly drawing $\vect{u}$'s from a standard normal. The KL term can be calculated analytically, which yields
\begin{equation}
\kl{q\br{\bm{\phi}}}{\prob{\bm{\phi}}} 
= \frac{\lambda}{2}\,\br{\norm{\bm{\phi}_m}^2 + \mathrm{sum}\br{\exp{\bm{\phi}_s}}}
 - \frac{1}{2}\mathrm{sum}\br{\bm{\phi}_s} + \mathrm{const}.
\end{equation}
The above optimization problem has been parameterized in such a way that $\bm{\phi}_m$ and $\bm{\phi}_s$ are unconstrained. Moreover, the derivatives of the variational lower bound with respect to $\bm{\phi}_m$ and $\bm{\phi}_s$  can be easily calculated with backpropagation after stochastic approximations to the expectations have been made. This allows the use of any off-the-shelf gradient-based stochastic optimizer. In our experiments, we implemented MDN-SVI in Theano \citep{Theano:2016}, which automatically calculates derivatives with backpropagation, and used Adam \citep{Kingma:2014} for stochastic maximization of the variational lower bound.

An important practical detail for stochastically approximating the expectation terms is the \emph{local reparameterization trick} \citep{Kingma:2015}, which leverages the layered feedforward structure of the MDN\@. Consider any hidden or output unit in the MDN; if $a$ is its activation and $\vect{z}$ is the vector of its inputs, then the relationship between $a$ and $\vect{z}$ is always of the form
\begin{equation}
a = \vect{w}^T\vect{z} + b,
\end{equation}
where $\vect{w}$ and $b$ are the weights and bias respectively associated with this unit. As we have seen, in the SVI framework these weights and biases are Gaussian random variables with means $\vect{w}_m$ and $b_m$, and log variances $\vect{w}_s$ and $b_s$. It is easy to see that this induces a Gaussian distribution over activation $a$, whose mean $a_m$ and variance $\exp{a_s}$ is given by
\begin{equation}
a_m = \vect{w}_m^T\vect{z} + b_m
\quad\text{ and }\quad
\exp{a_s} = \br{\exp{\vect{w}_s}}^T\br{\vect{z}\odot\vect{z}} + \exp{b_s},
\end{equation}
where $\odot$ denotes elementwise multiplication. Therefore, randomly sampling $\vect{w}$ and $b$ in order to estimate the expectations and their gradients in the variational lower bound is equivalent to directly sampling $a$ from a Gaussian with the above mean and variance. This trick saves computation by reducing calls to the random number generator, but more importantly it reduces the variance of the stochastic approximation of the expectations (intuitively this is because less randomness is involved) and hence it makes stochastic optimization more stable and faster to converge.

\section{Effective sample size of ABC methods}
\label{sec:ess_abc}

Rejection ABC returns a set of \emph{independent} samples, MCMC-ABC returns a set of \emph{correlated} samples, and SMC-ABC returns a set of independent but \emph{weighted} samples. To make a fair comparison between them in terms of simulation cost, we quote the number of simulations per \emph{effective} sample, that is, the total number of simulations divided by the effective sample size of the returned set of samples.

Let $\set{\bm{\theta}_n}$ be a set of $N$ samples, not necessarily independent. The effective sample size $N_\mathrm{eff}$  is defined to be the number of equivalent \emph{independent} samples that would give an estimator of equal variance. For rejection ABC $N_\mathrm{eff}=N$, since all returned samples are independent.

Suppose that each sample is a vector of $D$ components. For MCMC-ABC, where samples come in the form of $D$ autocorrelated sequences, we estimated the effective sample size for component $d$ as
\begin{equation}
N_{\mathrm{eff},d} = \frac{N}{1 + 2\sum_{l=1}^{L_d}{r_{dl}}},
\end{equation}
where $r_{dl}$ is the autocorrelation coefficient of component $d$ at lag $l$, estimated from the samples. We calculated the summation up to lag $L_d$, which corresponds to the first autocorrelation coefficient that is equal to $0$. Then we took the effective sample size $N_{\mathrm{eff}}$ to be the minimum $N_{\mathrm{eff},d}$ across components. For a more general discussion on estimating autocorrelation time (which is equal to $\nicefrac{N}{N_{\mathrm{eff}}}$ and thus equivalent to effective sample size) see \citet{Thompson:2010}.

For SMC-ABC, each sample is independent but comes with a corresponding non-negative weight $w_n$. The weights have to sum to one, that is $\sum_n{w_n}=1$. We estimated the effective sample size by
\begin{equation}
N_\mathrm{eff} = \frac{1}{\sum_n{w_n^2}}.
\end{equation}
It is easy to see that if $w_n = \nicefrac{1}{N}$ for all $n$ then $N_\mathrm{eff}=N$, and if all weights but one are $0$ then $N_\mathrm{eff}=1$. For a discussion regarding the above estimate see \citet{Nowozin:2015}.

\section{Setup for the Lotka--Volterra experiment}
\label{sec:lv_setup}

The Lotka--Volterra model \citep{Wilkinson:2011} is a stochastic model that was developed to describe the time evolution of a population of predators interacting with a population of prey. Let $X$ be the number of predators and $Y$ be the number of prey. The model asserts that the following four reactions can take place, with corresponding rates:
\begin{enumerate}[label=(\roman*)]
\item A predator may be born, with rate $\theta_1XY$, increasing $X$ by one.
\item A predator may die, with rate $\theta_2X$, decreasing $X$ by one.
\item A prey may be born, with rate $\theta_3Y$, increasing $Y$ by one.
\item A prey may be eaten by a predator, with rate $\theta_4XY$, decreasing $Y$ by one.
\end{enumerate}
Given initial populations $X$ and $Y$, the above model can be simulated using Gillespie's algorithm \citep{Gillespie:1977}, as follows:
\begin{enumerate}[label=(\roman*)]
\item Draw the time to next reaction from an exponential distribution with rate equal to the total rate $\theta_1XY + \theta_2X + \theta_3Y + \theta_4XY$.\label{alg:gillespie:step_1}
\item Select a reaction at random, with probability proportional to its rate.
\item Simulate the reaction, and go to step~\ref*{alg:gillespie:step_1}.
\end{enumerate}
In our experiments, each simulation started with initial populations $X=50$ and $Y=100$, and took place for a total of $30$ time units. We recorded the values of $X$ and $Y$ after every $0.2$ time units, resulting in two time series of $151$ values each.

\begin{figure}[t]
\def\imwidth{0.45\textwidth}
\centering
\subfloat{
\includegraphics[width=\imwidth]{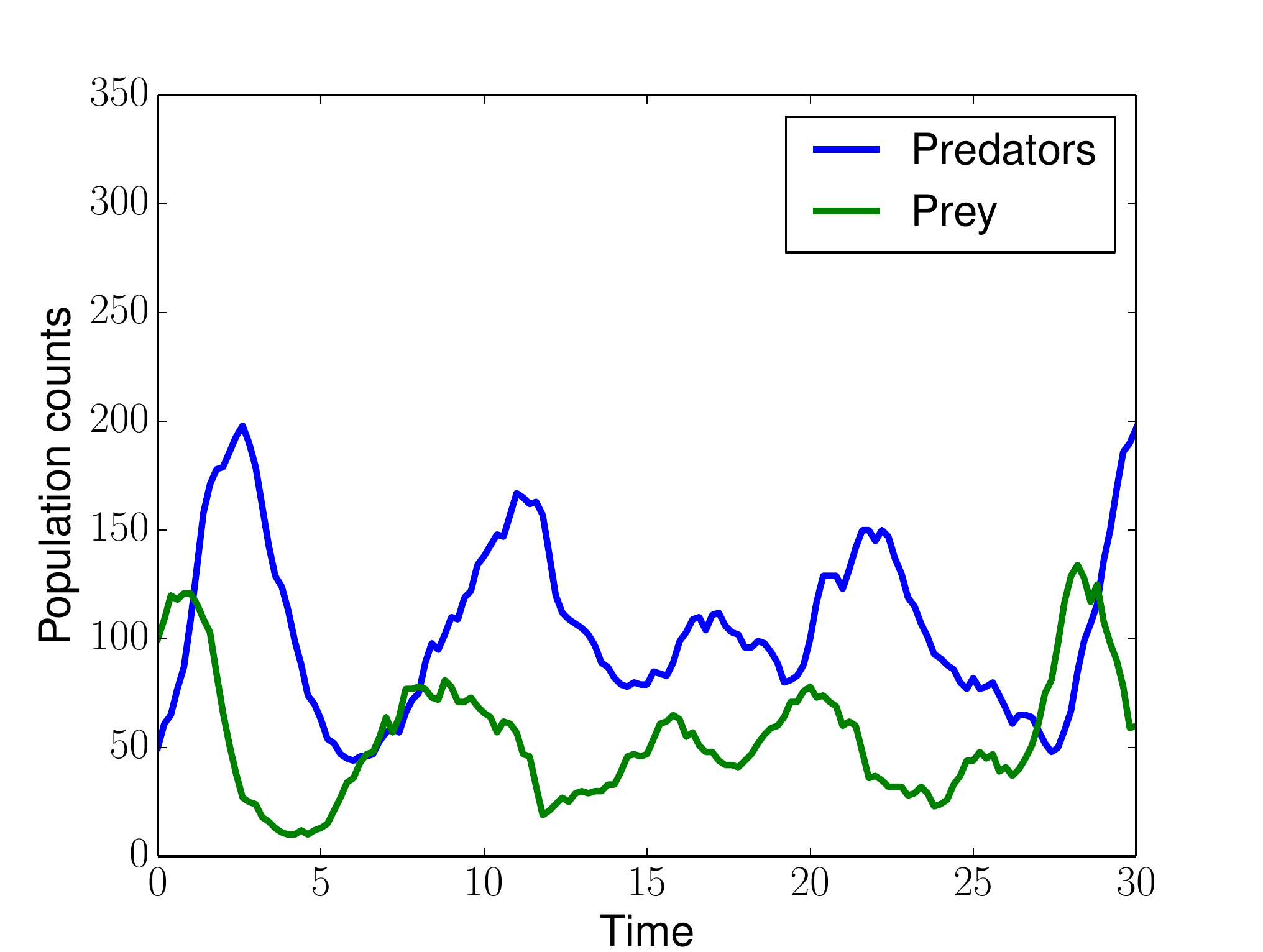}}
\subfloat{
\includegraphics[width=\imwidth]{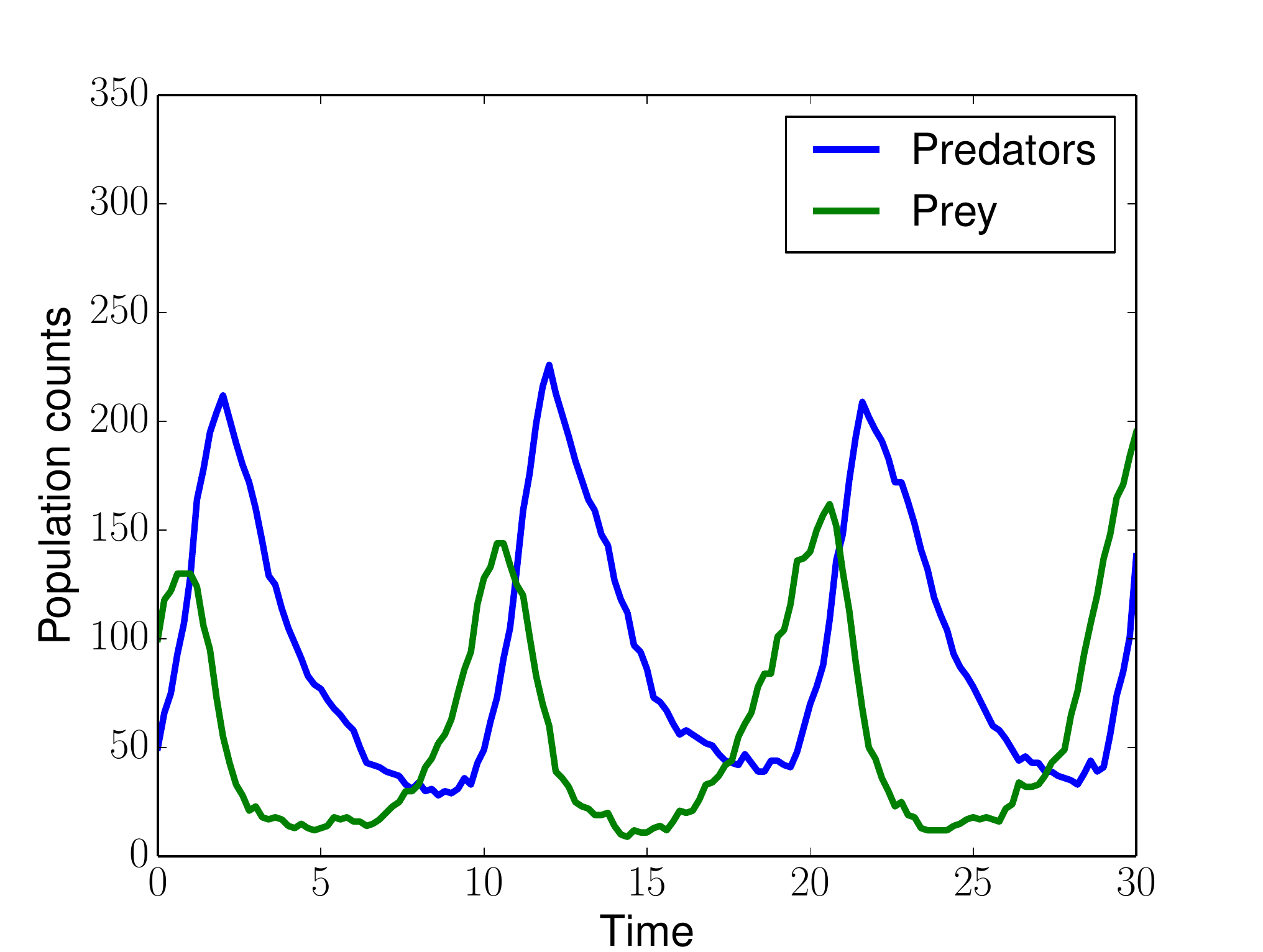}}\\
\subfloat{
\includegraphics[width=\imwidth]{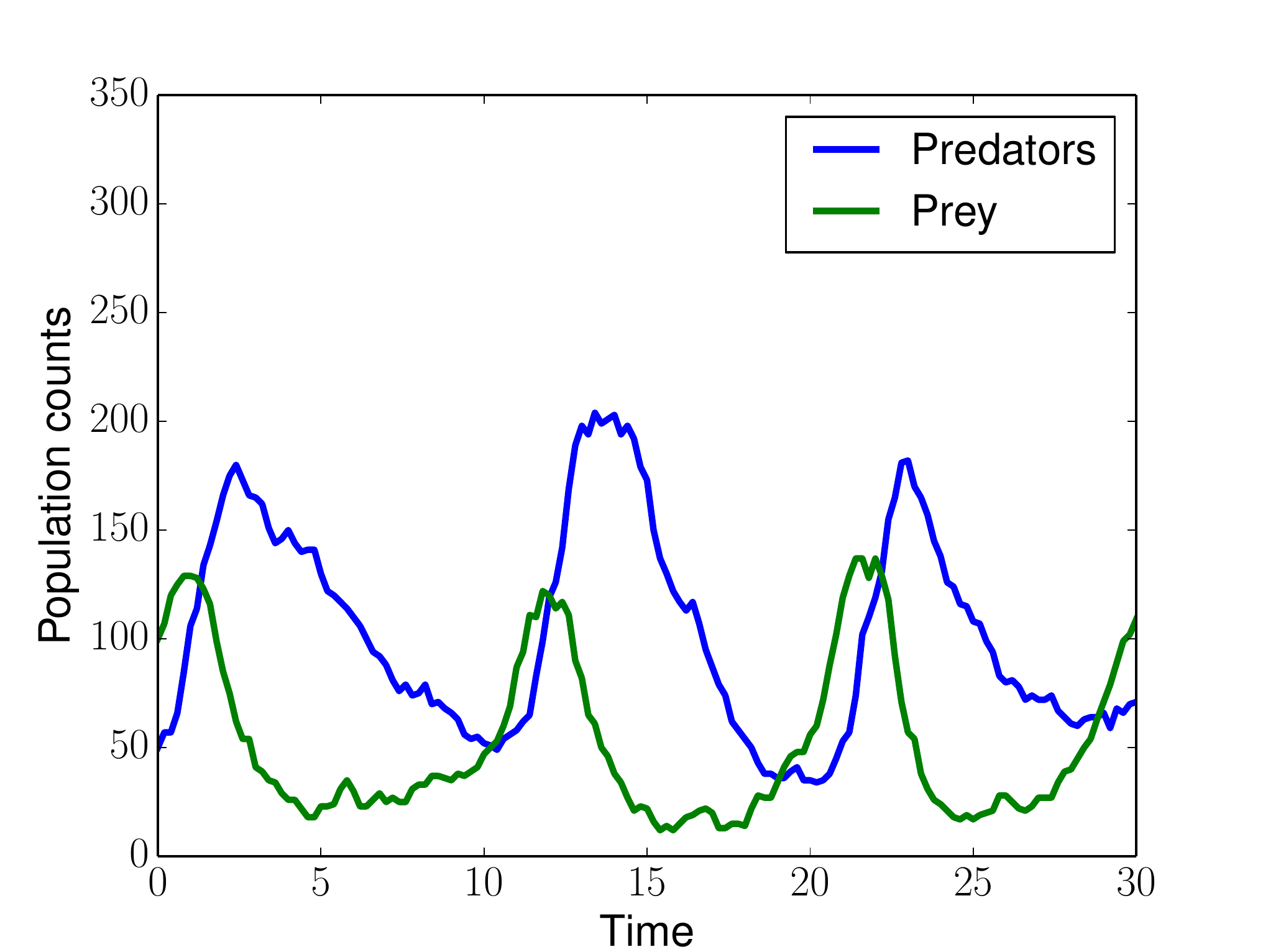}}
\subfloat{
\includegraphics[width=\imwidth]{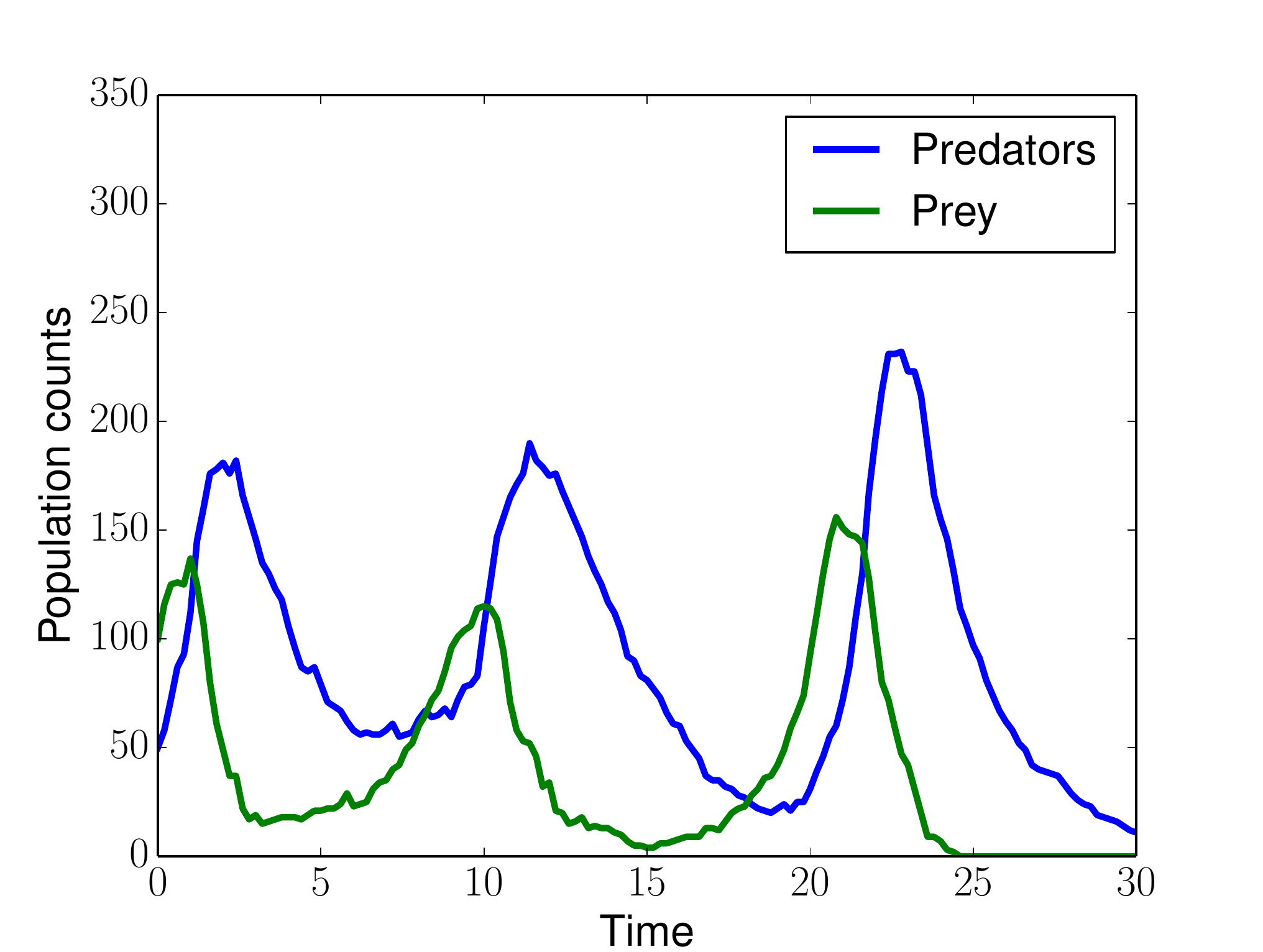}}
\caption{Typical oscillatory behaviour of predator/prey populations corresponding to four different simulations of the Lotka--Volterra model with parameter values $\theta_1 =0.01$, $\theta_2=0.5$, $\theta_3=1$, and $\theta_4=0.01$.}
\label{fig:lv_typical_sims}
\end{figure}

Data $\vect{x}$ was taken to be the following set of $9$ statistics calculated from the time series:
\begin{enumerate}[label=(\roman*)]
\item The mean of each time series.
\item The log variance of each time series.
\item The autocorrelation coefficient of each time series at lag $1$ and lag $2$.
\item The cross-correlation coefficient between the two time series.
\end{enumerate}
Since the above statistics have potentially very different scales, we normalized them on the basis of a pilot run. That is, we performed a pilot run of $1000$ simulations, calculated and stored the mean and standard deviation of each statistic across pilot simulations, and used them in all subsequent simulations to normalize each statistic by subtracting the pilot mean and dividing by the pilot standard deviation. This choice of statistics and normalization process was taken from \citet{Wilkinson:2013}.

From our experience with the model we observed that typical evolutions of the predator/prey populations for randomly selected parameters $\bm{\theta}=\br{\theta_1,\theta_2,\theta_3,\theta_4}$ include (a) the predators quickly eating all the prey and then slowly decaying exponentially, or (b) the predators quickly dying out and then the prey growing exponentially. However, for certain carefully tuned values of $\bm{\theta}$, the two populations exhibit an oscillatory behaviour, typical of natural ecological systems. In order to generate observations $\vect{x}_o$ for our experimental setup, we set the parameters to
\begin{equation}
\theta_1 =0.01,\quad\theta_2=0.5,\quad\theta_3=1,\quad\theta_4=0.01
\end{equation}
and simulated the model to generate $\vect{x}_o$. We carefully chose parameter values that give rise to oscillatory behaviour (see Figure~\ref{fig:lv_typical_sims} for typical examples of population evolution corresponding to the above parameters). Since only a small subset of parameters give rise to such oscillatory behaviour, the posterior $\prob{\bm{\theta}\g\vect{x}=\vect{x}_o}$ is expected to be tightly peaked around the true parameter values. We tested our algorithms by evaluating how well (in terms of assigned log probability) each algorithm retrieves the true parameters.

Finally, we took the prior over $\bm{\theta}$ to be uniform in the log domain. That is, the prior was taken to be
\begin{equation}
\prob{\log{\bm{\theta}}} \propto \prod_{i=1}^4{\uniform{\log{\theta_i}}{\log{\theta}_\alpha}{\log{\theta}_\beta}},
\end{equation}
where $\log{\theta}_\alpha=-5$ and $\log{\theta}_\beta=2$, which of course includes the true parameters. All our inferences where done in the log domain.

\section{Setup for the M/G/1 experiment}
\label{sec:mg1_setup}

The M/G/1 queue model \citep{Shestopaloff:2014} is a statistical model that describes how a single server processes a queue formed by a set of continuously arriving jobs. Let $I$ be the total number of jobs to be processed, $s_i$ be the time the server takes to process job $i$, $v_i$ be the time that job $i$ entered the queue, and $d_i$ be the time that job $i$ left the queue (i.e.~the time when the server finished processing it). The M/G/1 queue model asserts that for each job $i$ we have
\begin{align}
s_i &\sim \mathcal{U}\br{\theta_1,\theta_2} \\
v_i - v_{i-1} &\sim \mathrm{Exp}\br{\theta_3} \\
d_i - d_{i-1} &= s_i + \max{\br{0, v_i-d_{i-1}}}.
\end{align}
In the above equations, $\mathcal{U}\br{\theta_1,\theta_2}$ denotes a uniform distribution in the range $\left[\theta_1,\theta_2\right]$, $\mathrm{Exp}\br{\theta_3}$ denotes an exponential distribution with rate $\theta_3$, and $v_0=d_0=0$. In our experiments we used a total of $I=50$ jobs.

The goal is to infer parameters $\bm{\theta}=\br{\theta_1, \theta_2, \theta_3}$ if the only knowledge is a set of percentiles of the empirical distribution of the interdeparture times $d_i - d_{i-1}$ for $i=1,\ldots,I$. In our experiments we used $5$ equally spaced percentiles. That is, given a set of $I$ interdeparture times $d_i - d_{i-1}$, we took $\vect{x}$ to be the $0$th, $25$th, $50$th, $75$th and $100$th percentiles of the set of interdeparture times. Note that the $0$th and $100$th percentiles correspond to the minimum and maximum element in the set.

Since different percentiles can have different scales and strong correlations between them, we whitened the data on the basis of a pilot run. That is, we performed $100$K pilot simulations, and recorded the mean vector and covariance matrix of the resulting percentiles. For each subsequent simulation, we calculated $\vect{x}$ from resulting percentiles by subtracting the mean vector and decorrelating and normalizing with the covariance matrix.

To generate observed data $\vect{x}_o$, we set the parameters to the following values
\begin{equation}
\theta_1=1,\quad\theta_2=5,\quad\theta_3=0.2
\end{equation}
and simulated the model to get $\vect{x}_o$. We evaluated inference algorithms by how well the true parameter values were retrieved, as measured by log probability under computed posteriors. Finally, the prior probability of the parameters was taken to be
\begin{align}
\theta_1 &\sim \mathcal{U}\br{0,10} \\
\theta_2-\theta_1 &\sim \mathcal{U}\br{0,10} \\
\theta_3 &\sim \mathcal{U}\br{0,\nicefrac{1}{3}},
\end{align}
which is uniform, albeit not axis-aligned, and of course includes the true parameters.

\small{
\bibliography{references}
}

\end{document}